\documentclass{article}

\usepackage{arxiv}

\usepackage[utf8]{inputenc} 
\usepackage[T1]{fontenc}    
\usepackage[hyphens]{url}            
\usepackage{booktabs}       
\usepackage{amsfonts}       
\usepackage{nicefrac}       
\usepackage{microtype}      
\usepackage{lipsum}
\usepackage{amsmath}
\usepackage{amssymb}
\usepackage{bm}
\usepackage{bbold}
\usepackage{float}
\usepackage{graphicx}
\usepackage{subcaption}
\usepackage{multirow}
\usepackage{makecell}
\usepackage{blindtext}
\usepackage{tabularx}
\usepackage{xltabular}
\newcolumntype{C}{>{\centering\arraybackslash}X}
\newcolumntype{L}{>{\flushleft\arraybackslash}X}
\usepackage{breakcites}
\usepackage{afterpage}
\usepackage[inline]{enumitem}

\newlist{inlinelist}{enumerate*}{1}
\setlist[inlinelist]{label=(\roman*)}

\usepackage{siunitx}
\usepackage{caption}
\usepackage[dvipsnames, table]{xcolor}
\definecolor{darkgreen}{HTML}{4F7942}

\usepackage[hyperindex,breaklinks]{hyperref}       
\newcommand\myshade{85}
\colorlet{mylinkcolor}{violet}
\colorlet{mycitecolor}{Tan}
\colorlet{myurlcolor}{Aquamarine}

\hypersetup{
  linkcolor  = mylinkcolor!\myshade!black,
  citecolor  = mycitecolor!\myshade!black,
  urlcolor   = myurlcolor!\myshade!black,
  colorlinks = true,
}

\usepackage{tikz}
\usepackage{pgfplots}
\usepackage{pgfplotstable}                      
\usepgfplotslibrary{groupplots,dateplot}
\pgfplotsset{compat=newest}

\usetikzlibrary{plotmarks}
\usetikzlibrary{backgrounds}
\usetikzlibrary{positioning}
\usetikzlibrary{shapes}
\usetikzlibrary{arrows.meta}
\usetikzlibrary{patterns}
\usetikzlibrary{decorations.pathreplacing}
\usetikzlibrary{calligraphy}
\usetikzlibrary{pgfplots.statistics}        

\pgfplotsset{
    bplot style/.style={
        boxplot = {
            average=auto,
            every average/.style={mark=*, thick, mark size=2pt}
        },
        draw=black,
        solid,
        thick,
        mark=x,
        every mark/.append style={
            fill=black,
            opacity=0.8
        },
    },
	discard if/.style 2 args={
        x filter/.code={
            \edef\tempa{\thisrow{#1}}
            \edef\tempb{#2}
            \ifx\tempa\tempb
                
            \fi
        }
    },
    discard if not/.style 2 args={
        x filter/.code={
            \edef\tempa{\thisrow{#1}}
            \edef\tempb{#2}
            \ifx\tempa\tempb
            \else
                
            \fi
        }
    }
}

\definecolor{lightgray}{RGB}{230,230,230}
\definecolor{darkgray}{RGB}{150, 150, 150}
\definecolor{lightred}{RGB}{239, 139, 128}
\definecolor{darkred}{RGB}{165, 0, 0}
\definecolor{brightred}{RGB}{239, 74, 56}
\definecolor{lightblue}{RGB}{158, 186, 211}
\definecolor{brightblue}{RGB}{31, 126, 211}
\definecolor{lightorange}{RGB}{252, 209, 153}
\definecolor{brightorange}{RGB}{252, 165, 52}
\definecolor{lightyellow}{RGB}{240, 240, 180}

\usepackage[semicolon, authoryear]{natbib}

\usepackage{etoolbox}
\usepackage{tcolorbox}

\makeatletter
\pretocmd{\NAT@citex}{%
  \let\NAT@hyper@\NAT@hyper@citex
  \def\NAT@postnote{#2}%
  \setcounter{NAT@total@cites}{0}%
  \setcounter{NAT@count@cites}{0}%
  \forcsvlist{\stepcounter{NAT@total@cites}\@gobble}{#3}}{}{}
\newcounter{NAT@total@cites}
\newcounter{NAT@count@cites}
\def\NAT@postnote{}

\def\NAT@hyper@citex#1{%
  \stepcounter{NAT@count@cites}%
  \hyper@natlinkstart{\@citeb\@extra@b@citeb}#1%
  \ifnumequal{\value{NAT@count@cites}}{\value{NAT@total@cites}}
    {\ifNAT@swa\else\if*\NAT@postnote*\else%
     \NAT@cmt\NAT@postnote\global\def\NAT@postnote{}\fi\fi}{}%
  \ifNAT@swa\else\if\relax\NAT@date\relax
  \else\NAT@@close\global\let\NAT@nm\@empty\fi\fi
  \hyper@natlinkend}
\renewcommand\hyper@natlinkbreak[2]{#1}

\patchcmd{\NAT@citex}
  {\ifNAT@swa\else\if*#2*\else\NAT@cmt#2\fi
   \if\relax\NAT@date\relax\else\NAT@@close\fi\fi}{}{}{}
\patchcmd{\NAT@citex}
  {\if\relax\NAT@date\relax\NAT@def@citea\else\NAT@def@citea@close\fi}
  {\if\relax\NAT@date\relax\NAT@def@citea\else\NAT@def@citea@space\fi}{}{}

\makeatother

\usepackage{algorithm}
\usepackage[noend]{algpseudocode}

\DeclareSymbolFontAlphabet{\amsmathbb}{AMSb}%

\newcommand{\midr}{\specialrule{\lightrulewidth}{2mm}{1mm}}
\newcommand{\botr}{\specialrule{\lightrulewidth}{2mm}{0mm}}

\newcommand{\obox}[1]{\footnotesize\tcbox[on line, boxsep=2pt,left=2pt,right=2pt,top=2pt,bottom=2pt,colback=lightorange, colframe=black, boxrule=1pt]{#1}}
\newcommand{\rbox}[1]{\footnotesize\tcbox[on line, boxsep=2pt,left=2pt,right=2pt,top=2pt,bottom=2pt,colback=lightred, colframe=black, boxrule=1pt]{#1}}
\newcommand{\bbox}[1]{\footnotesize\tcbox[on line, boxsep=2pt,left=2pt,right=2pt,top=2pt,bottom=2pt,colback=lightblue, colframe=black, boxrule=1pt]{#1}}

\newcommand{\gbox}[1]{\footnotesize\tcbox[on line, boxsep=2pt,left=2pt,right=2pt,top=2pt,bottom=2pt,colback=lightgray, colframe=black, boxrule=1pt]{#1}}

\usepackage[show]{chato-notes} 

\let\vec\bm

\newif\ifarXiv
\arXivtrue 

\title{ciDATGAN: Conditional inputs for tabular GANs}

\author{
  Gael Lederrey \\
  Transport and Mobility Laboratory, \\ 
  \'Ecole Polytechnique F\'ed\'erale de Lausanne, \\
  Lausanne, Switzerland \\
  \texttt{gael.lederrey@epfl.ch}\\
   \And
  Tim Hillel \\
  Transport and Mobility Laboratory, \\ 
  \'Ecole Polytechnique F\'ed\'erale de Lausanne, \\
  Lausanne, Switzerland \\
  \texttt{tim.hillel@epfl.ch}\\
   \And
  Michel Bierlaire \\
  Transport and Mobility Laboratory, \\ 
  \'Ecole Polytechnique F\'ed\'erale de Lausanne, \\
  Lausanne, Switzerland \\
  \texttt{michel.bierlaire@epfl.ch}\\
}

\begin{document}
\maketitle

\begin{abstract}
  Conditionality has become a core component for Generative Adversarial Networks (GANs) for generating synthetic images. GANs are usually using latent conditionality to control the generation process. However, tabular data only contains manifest variables. Thus, latent conditionality either restricts the generated data or does not produce sufficiently good results. Therefore, we propose a new methodology to include conditionality in tabular GANs inspired by image completion methods. This article presents ciDATGAN, an evolution of the Directed Acyclic Tabular GAN (DATGAN) that has already been shown to outperform state-of-the-art tabular GAN models. First, we show that the addition of conditional inputs does hinder the model's performance compared to its predecessor. Then, we demonstrate that ciDATGAN can be used to unbias datasets with the help of well-chosen conditional inputs. Finally, it shows that ciDATGAN can learn the logic behind the data and, thus, be used to complete large synthetic datasets using data from a smaller feeder dataset.
\end{abstract}

\vfill


\newpage
\tableofcontents
\newpage

\section{Introduction}
\label{sec_ch3:introduction}

Synthetic data presents new opportunities for modeling complex systems with recent advances in deep generative learning methodologies. GANs~\citep{goodfellow_generative_2014} are considered the most advanced state-of-the-art models to generate such synthetic data. While there is a large body of work on generating synthetic images~\citep{shin_gans-awesome-applications_2022, hindupur_gan_2022}, there is still a gap in the research for generating tabular data. Multiple iterations of GANs for tabular data~\citep{xu_synthesizing_2018, park_data_2018, lederrey_datgan_2022} have been proposed. However, only a few examples of conditional GANs for tabular data exists~\citep{xu_modeling_2019, zhao_ctab-gan_2021}, and they lack the advances made for GANs specialized in images. Conditionality in GANs consists in feeding latent information about the desired synthetic data to control the generation process. For example, this can be used when one wants to generate images of a specific subject or generate images matching a descriptive sentence.

However, the concept of latent information in tabular data is less obvious. For example, for socio-economic data, much of the information that would be considered latent for an image (such as the individual's age, gender, or nationality) is typically included in the data as manifest variables. In this context, the conditionality of tabular data based on high-level descriptive details can be considered more similar to tasks such as image completion, where GANs attempt to recover the masked part of an image using the information from the neighboring pixels. In this chapter, we propose a new type of conditionality based on "\emph{conditional inputs}" that is a mix between latent conditionality and data completion. The idea is to feed some original data variables to the generator as conditional inputs and let the GAN learn the correlations with the other variables during the training process. Then, we can feed it some unknown values for the conditional inputs for the sampling. This does not break the principle of GANs since the generator never sees any original data it is trying to generate. This type of conditionality, thus, has two main use-cases:
\begin{enumerate}
    \item Removing the bias in a dataset, \emph{e.g.} correcting a non-representative sample in a survey data (\emph{e.g.} older individuals in a web-based survey)
    \item Combining information from two datasets, \emph{e.g.} combining representative aggregate data (\emph{e.g.} aggregate control totals from census) with detailed small-sample data (\emph{e.g.} household travel surveys) to create representative and fully disaggregate synthetic populations. 
\end{enumerate}

The conditional inputs methodology is directly implemented in the DATGAN~\citep{lederrey_datgan_2022}. For ease of comprehension, we name this model ciDATGAN. The code for the model itself is available on Github at \url{https://github.com/glederrey/DATGAN}. In addition, a PyPI library is also available at \url{https://pypi.org/project/datgan/}. Finally, the code that has been used to generate the results for this chapter is also available on Github at \url{https://github.com/glederrey/ciDATGAN}.

\section{Literature review}
\label{sec_ch3:lit_rev}

GANs~\citep{goodfellow_generative_2014} are considered the state-of-the-art generative models for creating synthetic datasets. While the fundamental concepts remain unchanged, many different approaches have been used depending on the research context and application. For this literature review, we investigate three types of GANs: conditional GANs (Section~\ref{sec_ch3:cond_gan}), image completion GANs (Section~\ref{sec_ch3:img_gan}), and tabular GANs (Section~\ref{sec_ch3:tab_gan}).

\subsection{Conditional GANs}
\label{sec_ch3:cond_gan}

GANs are built to generate highly representative synthetic data. However, it is usually impossible to control the generation process once the model has been trained. Therefore, the concept of conditionality has been introduced to overcome this limitation. It consists in feeding latent information about the desired synthetic data, such as the label of an image or a description, to both the generator and the discriminator during the training process. Thus, once the model has been trained, this information can be fed to the generator to control its sampling procedure. In the literature, we find four main types of conditionality: 
\begin{inlinelist}
    \item {\bf Conditionality by concatenation}~\citep{mirza_conditional_2014, reed_generative_2016} which consists of concatenating a vector of conditionals with the noise input for the generator and with the input values for the discriminator;
    \item {\bf Conditionality using an auxiliary classifier}~\citep{odena_conditional_2017} which consists of using an auxiliary neural network to classify the synthetic data and add the results to the loss function;
    \item {\bf Conditionality with projection}~\citep{miyato_cgans_2018} which consists of including a projection between the conditionality and the features extracted from the images in the discriminator, thus allowing the discriminator to compute the similarity between the conditionality and the image; and finally 
    \item {\bf Conditional Batch Normalization (CBN)}~\citep{dumoulin_learned_2017, de_vries_modulating_2017} which consists of using the conditional vectors to modulate the activation functions in the generator. \cite{zhang_self-attention_2019, brock_large_2019} used this idea in recent GAN architecture by replacing the Batch Normalization layers in the generator with CBN layers.
\end{inlinelist}

While all these methods have shown different levels of success until now, they have always been developed around the generation of synthetic images. The conditionality is defined relative to either: \begin{inlinelist}
\item the subject of the image;
\item a sentence describing an image; or
\item another image in a different style.
\end{inlinelist}
Thus, conditionality in GANs corresponds to latent conditionality, \emph{i.e.} it is not directly part of an image. The complement to latent conditionality in GANs is image completion.

\subsection{Image completion GANs}
\label{sec_ch3:img_gan}

Image completion GANs consist in generating part of an image that has been hidden with specific or random masks. \cite{yeh_semantic_2017} proposed a DCGAN for semantic inpainting~\citep{pathak_context_2016}. This model aims at inferring large missing parts of an image based on its semantics. Building on this work, \cite{nazeri_edgeconnect_2019} propose the EdgeConnect GAN that uses information about edges in the image to complete it. \cite{zheng_pluralistic_2019} extended the methodology to produce multiple possible completion for incomplete images. \cite{chen_generative_2020} proposed the iGPT model based on the well-known GPT3~\citep{brown_language_2020} model for language. Their model translates the idea of semantics for text to images.

\subsection{Tabular GANs}
\label{sec_ch3:tab_gan}

Since tabular data fundamentally differs from images, specific GANs have been developed to generate such data. Multiple architectures have been used for tabular GANs, including those based on: 
\begin{inlinelist}
\item LSTM cells~\citep{xu_synthesizing_2018, lederrey_datgan_2022};
\item FCNNs~\citep{xu_modeling_2019}; or
\item CNNs~\citep{zhao_ctab-gan_2021}.
\end{inlinelist} While these different architectures provide different strengths for each model, \cite{lederrey_datgan_2022} shows that their model, the DATGAN, outperforms the state-of-the-art models for generating representative synthetic data (without conditionality).

CTGAN~\citep{xu_modeling_2019} introduces conditionality for tabular GANs using concatenation. While they show that the conditional vector is critical for imbalanced datasets, they do not provide specific results on the conditionality. \cite{zhao_ctab-gan_2021} also implement conditionality in their CTAB-GAN. However, they use conditionality with an auxiliary classifier and, thus, have to restrict the conditionality on a single variable. Nonetheless, they show that in the context of ML classification, such conditionality improves the performance of resulting models by a significant margin.

Both methods for conditionality in tabular GANs cited here implement conditionality based on latent variables, which has been applied in image generation. As such, they either highly restrict the use of the model (conditionality using an auxiliary classifier) or fail to generate accurate synthetic data (conditionality by concatenation). Typically for tabular data, the types of variables that are latent in images are instead included explicitly in a dataset as manifest variables. For instance, a survey will typically contain the respondent's age, gender, and socio-demographic details. As such, we identify a gap in the literature for conditionality of tabular data generation based on manifest variables, which could be used in applications such as synthetic populations and data combinations. For example, to create a large and complete synthetic population, one would need to combine information from multiple datasets, \emph{e.g.} census data, activity timelines, and/or travel diaries. In this chapter, we address this gap by creating a methodology that allows the user to be flexible on its use of conditionality and the generation process of the model by taking inspiration from image completion methods.

\section{Methodology}
\label{sec_ch3:methodo}

This section describes our proposed methodology for generating synthetic tabular data based on conditional inputs. First, we introduce the concept of \emph{conditional inputs}. It takes inspiration from image completion, in contrast to latent conditionality. The overall approach is to train a model using complete data to generate synthetic observations conditional on one or more manifest variables. This type of conditionality can, thus, be used for either removing the bias in a dataset using unbiased variables as conditional inputs or combining information from two datasets using variables from a second dataset as conditional inputs for the sampling phase. 

\begin{figure}[!htb]
    \centering
    \begin{tikzpicture}[every text node part/.style={align=center}]

\makeatletter
\tikzset{
    database top segment style/.style={draw},
    database middle segment style/.style={draw},
    database bottom segment style/.style={draw},
    database/.style={
        path picture={
            \path [database bottom segment style]
                (-\db@r,-0.5*\db@sh) 
                -- ++(0,-1*\db@sh) 
                arc [start angle=180, end angle=360,
                    x radius=\db@r, y radius=\db@ar*\db@r]
                -- ++(0,1*\db@sh)
                arc [start angle=360, end angle=180,
                    x radius=\db@r, y radius=\db@ar*\db@r];
            \path [database middle segment style]
                (-\db@r,0.5*\db@sh) 
                -- ++(0,-1*\db@sh) 
                arc [start angle=180, end angle=360,
                    x radius=\db@r, y radius=\db@ar*\db@r]
                -- ++(0,1*\db@sh)
                arc [start angle=360, end angle=180,
                    x radius=\db@r, y radius=\db@ar*\db@r];
            \path [database top segment style]
                (-\db@r,1.5*\db@sh) 
                -- ++(0,-1*\db@sh) 
                arc [start angle=180, end angle=360,
                    x radius=\db@r, y radius=\db@ar*\db@r]
                -- ++(0,1*\db@sh)
                arc [start angle=360, end angle=180,
                    x radius=\db@r, y radius=\db@ar*\db@r];
            \path [database top segment style]
                (0, 1.5*\db@sh) circle [x radius=\db@r, y radius=\db@ar*\db@r];
        },
        minimum width=2*\db@r + \pgflinewidth,
        minimum height=3*\db@sh + 2*\db@ar*\db@r + \pgflinewidth,
    },
    database segment height/.store in=\db@sh,
    database radius/.store in=\db@r,
    database aspect ratio/.store in=\db@ar,
    database segment height=0.1cm,
    database radius=0.25cm,
    database aspect ratio=0.35,
    database top segment/.style={
        database top segment style/.append style={#1}},
    database middle segment/.style={
        database middle segment style/.append style={#1}},
    database bottom segment/.style={
        database bottom segment style/.append style={#1}}
}
\makeatother

\makeatletter
\tikzset{
    smalldb top segment style/.style={draw},
    smalldb middle segment style/.style={draw},
    smalldb/.style={
        path picture={
            \path [smalldb middle segment style]
                (-\db@r,-0*\db@sh) 
                -- ++(0,-1*\db@sh) 
                arc [start angle=180, end angle=360,
                    x radius=\db@r, y radius=\db@ar*\db@r]
                -- ++(0,1*\db@sh)
                arc [start angle=360, end angle=180,
                    x radius=\db@r, y radius=\db@ar*\db@r];
            \path [smalldb top segment style]
                (-\db@r,1*\db@sh) 
                -- ++(0,-1*\db@sh) 
                arc [start angle=180, end angle=360,
                    x radius=\db@r, y radius=\db@ar*\db@r]
                -- ++(0,1*\db@sh)
                arc [start angle=360, end angle=180,
                    x radius=\db@r, y radius=\db@ar*\db@r];
            \path [smalldb top segment style]
                (0, 1*\db@sh) circle [x radius=\db@r, y radius=\db@ar*\db@r];
        },
        minimum width=2*\db@r + \pgflinewidth,
        minimum height=2*\db@sh + 2*\db@ar*\db@r + \pgflinewidth,
    },
    smalldb segment height/.store in=\db@sh,
    smalldb radius/.store in=\db@r,
    smalldb aspect ratio/.store in=\db@ar,
    smalldb segment height=0.1cm,
    smalldb radius=0.25cm,
    smalldb aspect ratio=0.35,
    smalldb top segment/.style={
        smalldb top segment style/.append style={#1}},
    smalldb middle segment/.style={
        smalldb middle segment style/.append style={#1}},
}
\makeatother

\makeatletter
\tikzset{
    minidb top segment style/.style={draw},
    minidb/.style={
        path picture={
            \path [minidb top segment style]
                (-\db@r,0.5*\db@sh) 
                -- ++(0,-1*\db@sh) 
                arc [start angle=180, end angle=360,
                    x radius=\db@r, y radius=\db@ar*\db@r]
                -- ++(0,1*\db@sh)
                arc [start angle=360, end angle=180,
                    x radius=\db@r, y radius=\db@ar*\db@r];
            \path [minidb top segment style]
                (0, 0.5*\db@sh) circle [x radius=\db@r, y radius=\db@ar*\db@r];
        },
        minimum width=2*\db@r + \pgflinewidth,
        minimum height=\db@sh + 2*\db@ar*\db@r + \pgflinewidth,
    },
    minidb segment height/.store in=\db@sh,
    minidb radius/.store in=\db@r,
    minidb aspect ratio/.store in=\db@ar,
    minidb segment height=0.1cm,
    minidb radius=0.25cm,
    minidb aspect ratio=0.35,
    minidb top segment/.style={
        minidb top segment style/.append style={#1}},
}
\makeatother

\makeatletter
\tikzset{ loop/.style={ 
        draw,
        chamfered rectangle,
        chamfered rectangle xsep=2cm
    }
}
\makeatother

\draw[fill=lightgray] (0,0) -- (8,0) -- (8,1) -- (0,1) -- cycle;
\draw[fill=lightorange] (0,0) -- (-2,0) -- (-2,1) -- (0,1) -- cycle;

\draw[fill=lightorange] (0,-1) -- (-2,-1) -- (-2,  -5) -- (0, -5) -- cycle;
\draw[fill=lightgray, dashed] (0,-1) -- (8,-1) -- (8,  -5) -- (0, -5) -- cycle;
\draw[fill=gray] (-2,-1) -- (-3,-1) -- (-3,  -5) -- (-2, -5) -- cycle;

\draw[-{Triangle[scale=1.5]}, line width=1] (-3.5, 1.5) -- (0, 1.5);
\node[above] at (-1.75, 1.5) {columns};
\draw[-{Triangle[scale=1.5]}, line width=1] (-3.5, 1.5) -- (-3.5, -2);
\node[above, rotate=90] at (-3.5, -0.25) {rows};

\draw[-{Triangle[scale=1.5]}, line width=1] (8, 0.5) to[out=0,in=0] node[above, rotate=-90] {generated} (8, -3);

\draw [line width=1, decorate, decoration = {brace, raise=2pt, amplitude=7pt}] (-2,1) --  (8,1);
\node at (3, 1.6) {Feeder dataset ($\bf T_F$)};

\draw [line width=1, decorate, decoration = {brace, raise=2pt, amplitude=7pt, mirror}] (-3, -5) --  (0, -5);
\node at (-1.5,  -5.6) {Distributor dataset ($\bf T_D$)};

\draw [line width=1, decorate, decoration = {brace, raise=2pt, amplitude=5pt, mirror}] (-2,0) --  (0,0);
\draw [line width=1, decorate, decoration = {brace, raise=2pt, amplitude=5pt}] (-2,-1) --  (0,-1);
\node at (-1, -0.5) {Common variables};

\node at (-1, 0.5) {$\bf T^{ci}_F$};
\node at (4, 0.5) {$\bf T^{c}_F$};
\node at (-1, -3) {$\bf T^{ci}_D$};
\node at (4, -3) {$\bf T^{c, synth}_{F\rightarrow D}$};
\node at (-2.5, -3) {$\bf T^{c}_D$};

\end{tikzpicture}
    \caption{Representation of dataset completion in tabular data.}
    \label{fig_ch3:CI}
\end{figure}
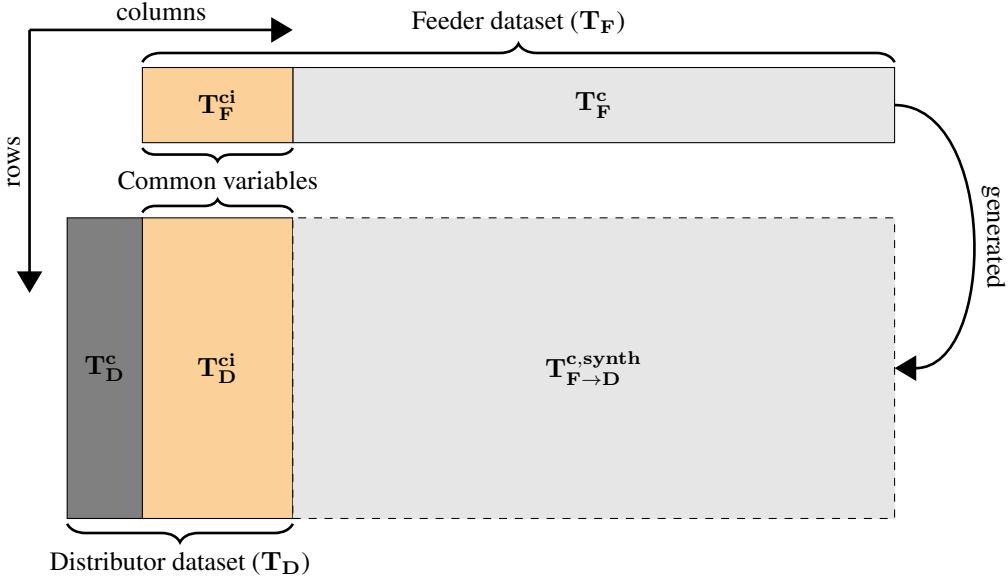

For example, datasets may be either high-detail small samples (with many variables and few rows) or low-detail large samples (with few rows but many variables). For example, in the case of population synthesis, one could have access to census data covering the whole population with high-level socio-economic characteristics and a more detailed dataset describing detailed mobility characteristics for a much smaller population sample. Thus, a model which learns how to generate the detailed mobility variables conditional on the high-level socio-economic census variables could be used to enrich the census population with these variables, thus creating a detailed synthetic dataset covering the whole of the population. We illustrate this example in Figure~\ref{fig_ch3:CI}. The dataset with many variables and fewer rows is named the feeder data ($\bf T_F$), and the dataset with low details but a large number of rows is named the distributor dataset ($\bf T_D$). As shown in Figure~\ref{fig_ch3:CI}, this methodology aims to learn part of the feeder dataset and generate synthetic data to complete the distributor dataset. It is done using common variables between the two datasets that we call conditional inputs.

Formally, the feeder dataset $\bf T_F$ contains $N_{\bf F}$ variables ($\vec{v}^{\bf F}_{i}$ for $i=1,\ldots,N_{\bf F}$) and the distributor dataset $\bf T_D$ contains $N_{\bf D}$ columns with $N_{\bf F} > N_{\bf D}$. To complete the distributor dataset with the information from the feeder dataset, we ensure that they contain $N_c \leq N_{\bf F}$ common variables. We designate these subsets as $\bf T^{ci}_F$ for the feeder dataset and $\bf T^{ci}_D$ for the distributor dataset. These common variables are the variables used as conditional inputs. The model has to learn the logic to generate the complementary variables in the feeder dataset ${\bf T^c_F} = {\bf T_F} \smallsetminus {\bf T^{ci}_F}$. In the sampling phase, it can complete the distributor dataset by generating the completementary variables of the feeder dataset $\bf T^{c,synth}_{F\rightarrow D}$ using the values of the common variables in the distributor dataset $\bf T^{ci}_D$ as conditional inputs. The final dataset is thus comprised of the distributor dataset $\bf T_D$ and the generated data $\bf T^{c,synth}_{F\rightarrow D}$. 

\subsection{ciDATGAN}
\label{sec_ch3:cidatgan}

\begin{figure}[!htb]
    \centering
    \begin{tikzpicture}[every text node part/.style={align=center}]

\makeatletter
\tikzset{
    database top segment style/.style={draw},
    database middle segment style/.style={draw},
    database bottom segment style/.style={draw},
    database/.style={
        path picture={
            \path [database bottom segment style]
                (-\db@r,-0.5*\db@sh) 
                -- ++(0,-1*\db@sh) 
                arc [start angle=180, end angle=360,
                    x radius=\db@r, y radius=\db@ar*\db@r]
                -- ++(0,1*\db@sh)
                arc [start angle=360, end angle=180,
                    x radius=\db@r, y radius=\db@ar*\db@r];
            \path [database middle segment style]
                (-\db@r,0.5*\db@sh) 
                -- ++(0,-1*\db@sh) 
                arc [start angle=180, end angle=360,
                    x radius=\db@r, y radius=\db@ar*\db@r]
                -- ++(0,1*\db@sh)
                arc [start angle=360, end angle=180,
                    x radius=\db@r, y radius=\db@ar*\db@r];
            \path [database top segment style]
                (-\db@r,1.5*\db@sh) 
                -- ++(0,-1*\db@sh) 
                arc [start angle=180, end angle=360,
                    x radius=\db@r, y radius=\db@ar*\db@r]
                -- ++(0,1*\db@sh)
                arc [start angle=360, end angle=180,
                    x radius=\db@r, y radius=\db@ar*\db@r];
            \path [database top segment style]
                (0, 1.5*\db@sh) circle [x radius=\db@r, y radius=\db@ar*\db@r];
        },
        minimum width=2*\db@r + \pgflinewidth,
        minimum height=3*\db@sh + 2*\db@ar*\db@r + \pgflinewidth,
    },
    database segment height/.store in=\db@sh,
    database radius/.store in=\db@r,
    database aspect ratio/.store in=\db@ar,
    database segment height=0.1cm,
    database radius=0.25cm,
    database aspect ratio=0.35,
    database top segment/.style={
        database top segment style/.append style={#1}},
    database middle segment/.style={
        database middle segment style/.append style={#1}},
    database bottom segment/.style={
        database bottom segment style/.append style={#1}}
}
\makeatother

\makeatletter
\tikzset{
    smalldb top segment style/.style={draw},
    smalldb middle segment style/.style={draw},
    smalldb/.style={
        path picture={
            \path [smalldb middle segment style]
                (-\db@r,-0*\db@sh) 
                -- ++(0,-1*\db@sh) 
                arc [start angle=180, end angle=360,
                    x radius=\db@r, y radius=\db@ar*\db@r]
                -- ++(0,1*\db@sh)
                arc [start angle=360, end angle=180,
                    x radius=\db@r, y radius=\db@ar*\db@r];
            \path [smalldb top segment style]
                (-\db@r,1*\db@sh) 
                -- ++(0,-1*\db@sh) 
                arc [start angle=180, end angle=360,
                    x radius=\db@r, y radius=\db@ar*\db@r]
                -- ++(0,1*\db@sh)
                arc [start angle=360, end angle=180,
                    x radius=\db@r, y radius=\db@ar*\db@r];
            \path [smalldb top segment style]
                (0, 1*\db@sh) circle [x radius=\db@r, y radius=\db@ar*\db@r];
        },
        minimum width=2*\db@r + \pgflinewidth,
        minimum height=2*\db@sh + 2*\db@ar*\db@r + \pgflinewidth,
    },
    smalldb segment height/.store in=\db@sh,
    smalldb radius/.store in=\db@r,
    smalldb aspect ratio/.store in=\db@ar,
    smalldb segment height=0.1cm,
    smalldb radius=0.25cm,
    smalldb aspect ratio=0.35,
    smalldb top segment/.style={
        smalldb top segment style/.append style={#1}},
    smalldb middle segment/.style={
        smalldb middle segment style/.append style={#1}},
}
\makeatother

\makeatletter
\tikzset{
    minidb top segment style/.style={draw},
    minidb/.style={
        path picture={
            \path [minidb top segment style]
                (-\db@r,0.5*\db@sh) 
                -- ++(0,-1*\db@sh) 
                arc [start angle=180, end angle=360,
                    x radius=\db@r, y radius=\db@ar*\db@r]
                -- ++(0,1*\db@sh)
                arc [start angle=360, end angle=180,
                    x radius=\db@r, y radius=\db@ar*\db@r];
            \path [minidb top segment style]
                (0, 0.5*\db@sh) circle [x radius=\db@r, y radius=\db@ar*\db@r];
        },
        minimum width=2*\db@r + \pgflinewidth,
        minimum height=\db@sh + 2*\db@ar*\db@r + \pgflinewidth,
    },
    minidb segment height/.store in=\db@sh,
    minidb radius/.store in=\db@r,
    minidb aspect ratio/.store in=\db@ar,
    minidb segment height=0.1cm,
    minidb radius=0.25cm,
    minidb aspect ratio=0.35,
    minidb top segment/.style={
        minidb top segment style/.append style={#1}},
}
\makeatother

\makeatletter
\tikzset{ loop/.style={ 
        draw,
        chamfered rectangle,
        chamfered rectangle xsep=2cm
    }
}
\makeatother

\def\n{20}

\def\dy{2}
\def\dx{3.8}



\node[ellipse, fill=lightred, line width=1.5, draw=black, minimum width=2cm, minimum height=1.5cm] (A) at (0*\dx,0*\dy) {DAG};

\node[ellipse, fill=lightblue, line width=1.5, draw=black, minimum width=2cm, minimum height=1.5cm] (B) at (2.15*\dx,1.1*\dy) {Noise};

\node[rounded corners=10pt, fill=lightgray, line width=1.5, draw=black, minimum width=2.5cm, minimum height=1.5cm] (C) at (1.3*\dx,0) {Generator};

\node[rounded corners=10pt, fill=lightgray, line width=1.5, draw=black, minimum width=2.5cm, minimum height=1.5cm] (D) at (3*\dx,0) {Discriminator};

\node[database,database radius=1cm,database segment height=0.4cm, line width=1.5] (E) at (0*\dx,2*\dy) {};
\node at (0*\dx,2*\dy+0.6) {$\bf T_F$};

\node[smalldb,smalldb radius=1cm,smalldb segment height=0.4cm, line width=1.5, smalldb middle segment={fill=white}, smalldb top segment={fill=white}]  (F) at (2.15*\dx,-\dy) {};
\node at (2.15*\dx,-\dy+0.4) {\small $\bf T^{c,synth}_F$};

\node[minidb,minidb radius=1cm,minidb segment height=0.4cm, line width=1.5, minidb top segment={fill=lightorange}]  (H) at (1.3*\dx,1.7*\dy) {};
\node at (1.3*\dx,1.58*\dy+0.4) {\small $\bf T^{ci}_F$};

\node[smalldb,smalldb radius=1cm,smalldb segment height=0.4cm, line width=1.5, smalldb middle segment={fill=white}, smalldb top segment={fill=white}]  (G) at (1.3*\dx,2.25*\dy) {};
\node at (1.3*\dx,2.25*\dy+0.4) {\small $\bf T^c_F$};

\coordinate (C00) at (0.51*\dx, 2*\dy);
\coordinate (C0) at (0.75*\dx, 2*\dy);
\draw[-{Triangle[scale=1.5]}, line width=1] ([yshift=1pt]C00.east) -- ([yshift=1pt]C0.east) |- (G);
\draw[-{Triangle[scale=1.5]}, line width=1] ([yshift=-1pt]C00.east) -- ([yshift=-1pt]C0.east) |- (H);
\draw[line width=1] (E) -- (C00);

\draw[dashed, line width=1] ([xshift=-3pt]H.south) |- (0*\dx,1.1*\dy);
\draw[-{Triangle[scale=1.5]}, line width=1, dashed] (E.south) --  (A.north);

\coordinate (C1) at (1.3*\dx, 1.1*\dy);
\draw[-{Triangle[scale=1.5, color=white]}, line width=1] ([xshift=-1pt]H.south) -- ([xshift=-1pt]C.north);
\draw[-{Triangle[scale=1.5, color=white]}, line width=1] (B)  -- ([xshift=1pt]C1) -| ([xshift=1pt]C.north);
\draw[-{Triangle[scale=1.5]}, line width=1] ([yshift=1pt]C.north) -- (C.north);

\draw[-{Triangle[scale=1.5]}, line width=1] (C) |- (F);
\draw[-{Triangle[scale=1.5]}, line width=1] (F) -| (D);

\node[fill=white, draw=black, rounded corners=5pt, line width=1] at (1.3*\dx,-0.75*\dy) {output};
\node[fill=white, draw=black, rounded corners=5pt, line width=1] at (3*\dx,-0.75*\dy) {input};

\draw[-{Triangle[scale=1.5]}, line width=1] (G) -| (D);

\draw[-{Triangle[scale=1.5]}, line width=1, dashed] (A) -- (C);
\draw[-{Triangle[scale=1.5]}, line width=1, dashed] (D) -- (C);

\node[fill=white, draw=black, rounded corners=5pt, line width=1] at (1.3*\dx,0.75*\dy) {inputs};
\node[fill=white, draw=black, rounded corners=5pt, line width=1] at (0*\dx,1.1*\dy) {design};
\node[fill=white, draw=black, rounded corners=5pt, line width=1] at (3*\dx,0.75*\dy) {input};
\node[fill=white, draw=black, rounded corners=5pt, line width=1] at (0.51*\dx, 2*\dy) {split};
\node[fill=white, draw=black, rounded corners=5pt, line width=1] at (2.15*\dx, 0) {backpropagation};
\node[fill=white, draw=black, rounded corners=5pt, line width=1] at (0.57*\dx, 0) {structure};

\end{tikzpicture}
    \caption{Schema of the training process of ciDATGAN.}
    \label{fig_ch3:training}
\end{figure}
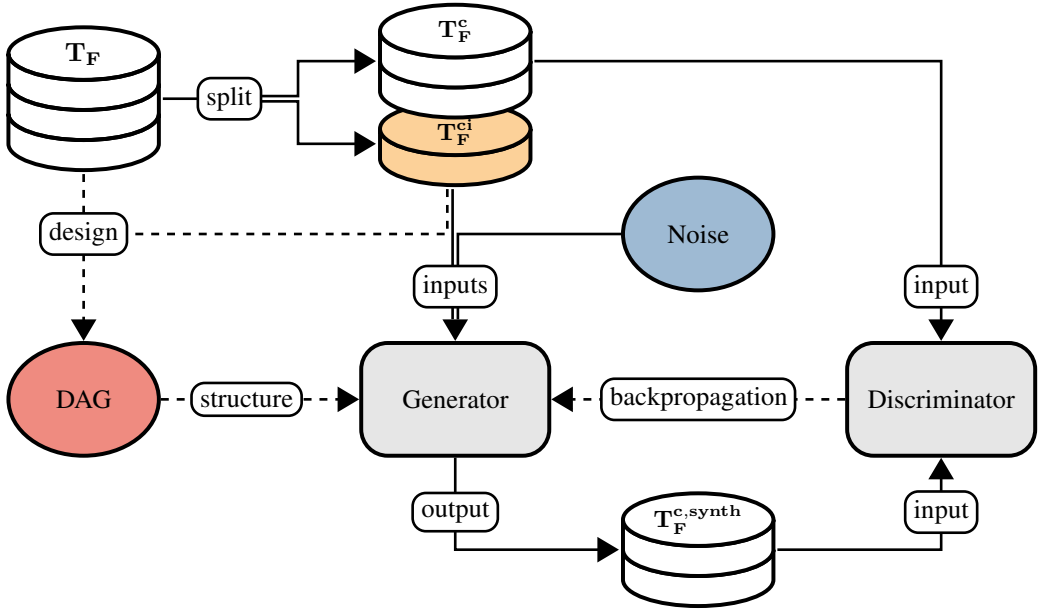

ciDATGAN is a direct extension of DATGAN~\citep{lederrey_datgan_2022}. The core methodology is the same for both models. Therefore, we give here a summary. The reader will find all the details in~\cite{lederrey_datgan_2022}. 

ciDATGAN uses LSTM cells~\citep{hochreiter_long_1997} in the generator to generate each variable, with a network structure based on a DAG provided by the user which specifies the causal links between variables in the feeder dataset. The variables in the data are encoded at discriminator input (and generator input for the conditional inputs) and decoded at generator output, with different encoding/decoding used for continuous and categorical variables. 

Figure~\ref{fig_ch3:training} shows how the different elements are combined to train ciDATGAN. First, the DAG is modified based on the selected variables in $\bf T_F^{ci}$. This modified DAG is then used to form the generator structure in an automated process. As for any GAN, the generator uses Gaussian noise as input. However, in ciDATGAN, we also include as input the conditional inputs $\bf T^{ci}_F$. Ultimately, the complementary variables $\bf T_F^{c}$ are used to train the discriminator against the generated data $\bf T_F^{c, synth}$. Thus, the training process remains similar to any existing GAN. 

Between DATGAN and ciDATGAN, there are two primary modifications to integrate the conditional inputs. First, the variables selected as conditional inputs must be considered as the generator's inputs. We, thus, impose each variable in $\bf T_F^{ci}$ to be a source node in the DAG. This modified DAG can be obtained through automated modification of an existing DAG.

Secondly, the conditional inputs can not be treated the same way as the generator's variables in $\bf T_F^c$. Indeed, these variables are generated using LSTM cells, similarly to DATGAN. We give here a summary of how these variables are generated. The LSTM cells follow an order provided by the linearization of the DAG. The cell $\text{LSTM}_t$ is associated with the variable $\vec{v}^{\bf F}_{t}$, in which the indices are ordered based on the DAG. Each cell takes as input the cell state of the previous variable in the DAG $\vec{C}_{t-1}$ and the input tensor $\vec{i}_t$. The input tensor is a concatenation of the following tensors:
\begin{itemize}
    \item $\vec{z}_t$: a tensor of Gaussian noise
    \item $\vec{f}_{t-1}$: the transformed output of the previous LSTM cell in the DAG. 
    \item $\vec{a}_t$: the attention vector used to retain information from previous ancestors that are not directly linked to the current cell in the DAG. It is defined as:
    \begin{equation}
        \vec{a}_t = \sum_{k\in\mathcal{A}(t)\setminus\mathcal{P}(t)} \frac{\exp\alpha^{(t)}_k}{\sum^{|\alpha^{(t)}|}_{j=1}\exp\alpha^{(t)}_j}\vec{f}_k 
    \end{equation}
    where $\mathcal{A}(t)\setminus\mathcal{P}(t)$ is the set of ancestors of the variable $\vec{v}^{\bf F}_t$ in the DAG in which we removed the direct predecessor(s), $\alpha^{(t)}$ is a learned attention weight vector, and $\vec{f}_k$ is the final output of the LSTM cell $\bf LSTM_k$.
\end{itemize}

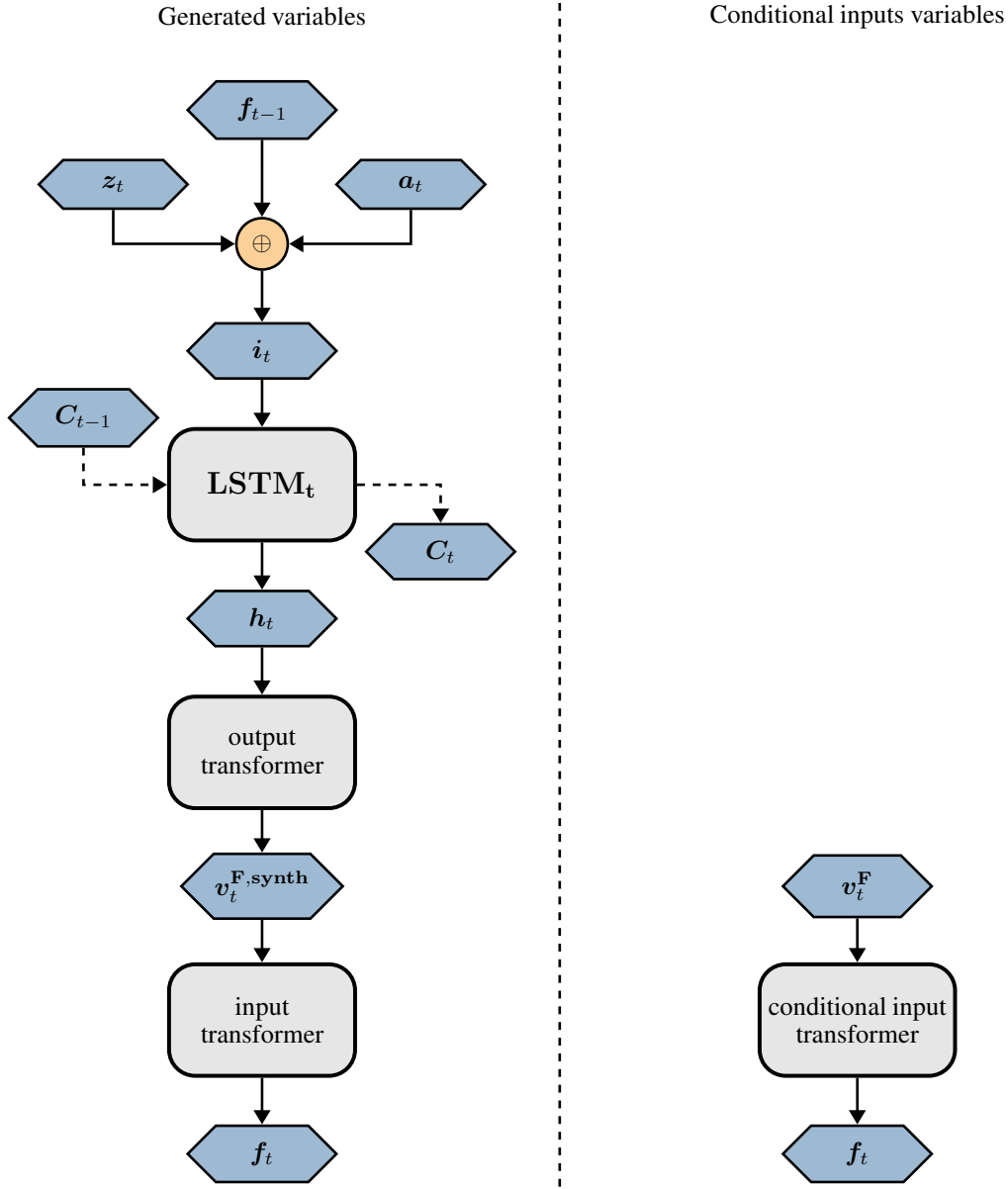
\begin{figure}[!htb]
    \centering
    \begin{tikzpicture}[every text node part/.style={align=center}]

\makeatletter
\tikzset{ loop/.style={ 
        draw,
        chamfered rectangle,
        chamfered rectangle xsep=2cm
    }
}
\makeatother

\def\n{20}

\def\dy{1.8}
\def\dx{4}

\node at (0, 3.5*\dy) {Generated variables};
\node at (2*\dx, 3.5*\dy) {Conditional inputs variables};

\node[rounded corners=10pt, fill=lightgray, line width=1.5, draw=black, minimum width=2.5cm, minimum height=1.5cm] (A) at (0,0) {\large $\mathbf{LSTM_t}$};

\node[loop, fill=lightblue, draw=black, line width=1, minimum width=2cm] (B) at (-0.6*\dx, 0.5*\dy) {$\vec{C}_{t-1}$};
\draw[-{Triangle[scale=1]}, line width=1, dashed] (B) |- (A);

\node[loop, fill=lightblue, draw=black, line width=1, minimum width=2cm] (P) at (0,\dy) {$\vec{i}_{t}$};
\draw[-{Triangle[scale=1]}, line width=1] (P) -- (A);

\node[circle, fill=lightorange, draw=black, line width=1] (C) at (0, 1.8*\dy) {$\oplus$};
\draw[-{Triangle[scale=1]}, line width=1] (C) -- (P);

\node[loop, fill=lightblue, draw=black, line width=1, minimum width=2cm] (D) at (0.6*\dx,-0.5*\dy) {$\vec{C}_{t}$};
\draw[-{Triangle[scale=1]}, line width=1, dashed] (A.east) -| (D);

\node[loop, fill=lightblue, draw=black, line width=1, minimum width=2cm] (E) at (0,-\dy) {$\vec{h}_{t}$};
\draw[-{Triangle[scale=1]}, line width=1] (A) -- (E);

\node[loop, fill=lightblue, draw=black, line width=1, minimum width=2cm] (F) at (-\dx/2,2.8*\dy-1) {$\vec{z}_{t}$};
\draw[-{Triangle[scale=1]}, line width=1] (F) |- (C);
\node[loop, fill=lightblue, draw=black, line width=1, minimum width=2cm] (G) at (0,2.8*\dy) {$\vec{f}_{t-1}$};
\draw[-{Triangle[scale=1]}, line width=1] (G) -- (C);
\node[loop, fill=lightblue, draw=black, line width=1, minimum width=2cm] (H) at (\dx/2,2.8*\dy-1) {$\vec{a}_{t}$};
\draw[-{Triangle[scale=1]}, line width=1] (H) |- (C);

\node[rounded corners=10pt, fill=lightgray, line width=1.5, draw=black, minimum width=2.5cm, minimum height=1.5cm] (I) at (0,-2*\dy) {output\\ transformer};
\draw[-{Triangle[scale=1]}, line width=1] (E) -- (I);

\node[rounded corners=10pt, fill=lightgray, line width=1.5, draw=black, minimum width=2.5cm, minimum height=1.5cm] (J) at (0,-4*\dy) {input\\ transformer};

\node[loop, fill=lightblue, draw=black, line width=1, minimum width=2cm] (K) at (0,-5*\dy) {$\vec{f}_{t}$};
\draw[-{Triangle[scale=1]}, line width=1] (J) -- (K);

\node[loop, fill=lightblue, draw=black, line width=1, minimum width=2cm] (L) at (0,-3*\dy) {$\vec{v}_t^{\bf F, synth}$};
\draw[-{Triangle[scale=1]}, line width=1] (I) -- (L);
\draw[-{Triangle[scale=1]}, line width=1] (L) -- (J);

\node[loop, fill=lightblue, draw=black, line width=1, minimum width=2cm] (M) at (2*\dx,-3*\dy) {$\vec{v}^{\bf F}_{t}$};

\node[rounded corners=10pt, fill=lightgray, line width=1.5, draw=black, minimum width=2.5cm, minimum height=1.5cm] (P) at (2*\dx,-4*\dy) {conditional input\\ transformer};

\node[loop, fill=lightblue, draw=black, line width=1, minimum width=2cm] (Q) at (2*\dx,-5*\dy) {$\vec{f}_{t}$};

\draw[-{Triangle[scale=1]}, line width=1] (M) -- (P);
\draw[-{Triangle[scale=1]}, line width=1] (P) -- (Q);

\draw[dashed, line width=1] (1*\dx, 3.6*\dy) -- (1*\dx, -5.3*\dy);

\node at (2.7*\dx, 0*\dy) {};

\node[loop, fill=white, draw=white, line width=1, minimum width=2cm] (B) at (2.7*\dx, 0.5*\dy) {};

\end{tikzpicture}
    \caption{Schematic representation of the variables in the generator. On the left, we show how the variables in ${\bf T^c}$ are generated using LSTM cells. On the right, we show how the variables in ${\bf T^{ci}}$ are transformed to match the elements on the left.}
    \label{fig_ch3:lstm}
\end{figure}

Each cell outputs two tensors: the new cell state $\vec{C}_t$ and the output of the cell $\vec{h}_t$. This output is then passed through a set of fully connected layers to get the synthetic values $\vec{v}_t^{\bf F, synth}$. Finally, since this synthetic tensor can have different dimensions depending on the encoding of the original columns $\vec{v}^{\bf F}_t$, it is passed through an input transformer to resize it to a common size between all variables. Figure~\ref{fig_ch3:lstm} shows the schematic representation of the generation of these synthetic variables. 

We only have access to the values themselves for the variables in $\bf T_F^{ci}$. However, the generator needs to have the two following values for each variable: the transformed output $\vec{f}_{t-1}$ of the direct ancestor and the direct output $\vec{h}_k$ of all the ancestors. Therefore, the variables in $\bf T_F^{ci}$ must be transformed accordingly. First, we use the same type of input transformer to get $\vec{f}_t$. However, we cannot get access to the LSTM output $\vec{h}_t$ for the variables in the conditional inputs $\bf T_F^{ci}$ since we do not use LSTM cells for these variables. Therefore, we transform the original value $\vec{v}_t$ using a Dense layer. The parameters in this additional dense layer are learned during the training process. This allows the model to use the conditional inputs in the attention vector. Figure~\ref{fig_ch3:lstm} shows how the variables $\vec{v}_t$ are transformed when they are considered as conditional inputs. 

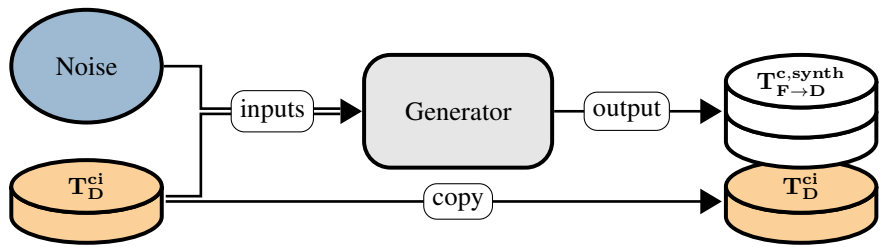
\begin{figure}[!htb]
    \centering
    \begin{tikzpicture}[every text node part/.style={align=center}]

\makeatletter
\tikzset{
    database top segment style/.style={draw},
    database middle segment style/.style={draw},
    database bottom segment style/.style={draw},
    database/.style={
        path picture={
            \path [database bottom segment style]
                (-\db@r,-0.5*\db@sh) 
                -- ++(0,-1*\db@sh) 
                arc [start angle=180, end angle=360,
                    x radius=\db@r, y radius=\db@ar*\db@r]
                -- ++(0,1*\db@sh)
                arc [start angle=360, end angle=180,
                    x radius=\db@r, y radius=\db@ar*\db@r];
            \path [database middle segment style]
                (-\db@r,0.5*\db@sh) 
                -- ++(0,-1*\db@sh) 
                arc [start angle=180, end angle=360,
                    x radius=\db@r, y radius=\db@ar*\db@r]
                -- ++(0,1*\db@sh)
                arc [start angle=360, end angle=180,
                    x radius=\db@r, y radius=\db@ar*\db@r];
            \path [database top segment style]
                (-\db@r,1.5*\db@sh) 
                -- ++(0,-1*\db@sh) 
                arc [start angle=180, end angle=360,
                    x radius=\db@r, y radius=\db@ar*\db@r]
                -- ++(0,1*\db@sh)
                arc [start angle=360, end angle=180,
                    x radius=\db@r, y radius=\db@ar*\db@r];
            \path [database top segment style]
                (0, 1.5*\db@sh) circle [x radius=\db@r, y radius=\db@ar*\db@r];
        },
        minimum width=2*\db@r + \pgflinewidth,
        minimum height=3*\db@sh + 2*\db@ar*\db@r + \pgflinewidth,
    },
    database segment height/.store in=\db@sh,
    database radius/.store in=\db@r,
    database aspect ratio/.store in=\db@ar,
    database segment height=0.1cm,
    database radius=0.25cm,
    database aspect ratio=0.35,
    database top segment/.style={
        database top segment style/.append style={#1}},
    database middle segment/.style={
        database middle segment style/.append style={#1}},
    database bottom segment/.style={
        database bottom segment style/.append style={#1}}
}
\makeatother

\makeatletter
\tikzset{
    smalldb top segment style/.style={draw},
    smalldb middle segment style/.style={draw},
    smalldb/.style={
        path picture={
            \path [smalldb middle segment style]
                (-\db@r,-0*\db@sh) 
                -- ++(0,-1*\db@sh) 
                arc [start angle=180, end angle=360,
                    x radius=\db@r, y radius=\db@ar*\db@r]
                -- ++(0,1*\db@sh)
                arc [start angle=360, end angle=180,
                    x radius=\db@r, y radius=\db@ar*\db@r];
            \path [smalldb top segment style]
                (-\db@r,1*\db@sh) 
                -- ++(0,-1*\db@sh) 
                arc [start angle=180, end angle=360,
                    x radius=\db@r, y radius=\db@ar*\db@r]
                -- ++(0,1*\db@sh)
                arc [start angle=360, end angle=180,
                    x radius=\db@r, y radius=\db@ar*\db@r];
            \path [smalldb top segment style]
                (0, 1*\db@sh) circle [x radius=\db@r, y radius=\db@ar*\db@r];
        },
        minimum width=2*\db@r + \pgflinewidth,
        minimum height=2*\db@sh + 2*\db@ar*\db@r + \pgflinewidth,
    },
    smalldb segment height/.store in=\db@sh,
    smalldb radius/.store in=\db@r,
    smalldb aspect ratio/.store in=\db@ar,
    smalldb segment height=0.1cm,
    smalldb radius=0.25cm,
    smalldb aspect ratio=0.35,
    smalldb top segment/.style={
        smalldb top segment style/.append style={#1}},
    smalldb middle segment/.style={
        smalldb middle segment style/.append style={#1}},
}
\makeatother

\makeatletter
\tikzset{
    minidb top segment style/.style={draw},
    minidb/.style={
        path picture={
            \path [minidb top segment style]
                (-\db@r,0.5*\db@sh) 
                -- ++(0,-1*\db@sh) 
                arc [start angle=180, end angle=360,
                    x radius=\db@r, y radius=\db@ar*\db@r]
                -- ++(0,1*\db@sh)
                arc [start angle=360, end angle=180,
                    x radius=\db@r, y radius=\db@ar*\db@r];
            \path [minidb top segment style]
                (0, 0.5*\db@sh) circle [x radius=\db@r, y radius=\db@ar*\db@r];
        },
        minimum width=2*\db@r + \pgflinewidth,
        minimum height=\db@sh + 2*\db@ar*\db@r + \pgflinewidth,
    },
    minidb segment height/.store in=\db@sh,
    minidb radius/.store in=\db@r,
    minidb aspect ratio/.store in=\db@ar,
    minidb segment height=0.1cm,
    minidb radius=0.25cm,
    minidb aspect ratio=0.35,
    minidb top segment/.style={
        minidb top segment style/.append style={#1}},
}
\makeatother

\makeatletter
\tikzset{ loop/.style={ 
        draw,
        chamfered rectangle,
        chamfered rectangle xsep=2cm
    }
}
\makeatother

\def\n{20}

\def\dy{3}
\def\dx{3.8}



\node[minidb,minidb radius=1cm,minidb segment height=0.4cm, line width=1.5, minidb top segment={fill=lightorange}]  (A) at (0*\dx,-0.4*\dy) {};
\node at (0*\dx,-0.33*\dy) {\small $\bf T^{ci}_D$};

\node[rounded corners=10pt, fill=lightgray, line width=1.5, draw=black, minimum width=2.5cm, minimum height=1.5cm] (B) at (1.3*\dx,0) {Generator};

\node[minidb,minidb radius=1cm,minidb segment height=0.4cm, line width=1.5, minidb top segment={fill=lightorange}]  (C) at (2.5*\dx,-0.4*\dy) {};
\node at (2.5*\dx,-0.33*\dy) {\small $\bf T^{ci}_D$};

\node[smalldb,smalldb radius=1cm,smalldb segment height=0.4cm, line width=1.5, smalldb middle segment={fill=white}, smalldb top segment={fill=white}]  (D) at (2.5*\dx,0*\dy) {};
\node at (2.5*\dx,0*\dy+0.4) {\small $\bf T^{c,synth}_{F\rightarrow D}$};

\node[ellipse, fill=lightblue, line width=1.5, draw=black, minimum width=2cm, minimum height=1.5cm] (E) at (0*\dx,0.2*\dy) {Noise};

\coordinate (C1) at (0.4*\dx, 0);

\draw[-{Triangle[scale=1.5, color=white]}, line width=1] ([yshift=2pt]A.east) -| ([yshift=-1pt]C1) -- ([yshift=-1pt]B.west);
\draw[-{Triangle[scale=1.5, color=white]}, line width=1] (E) -| ([yshift=1pt]C1) -- ([yshift=1pt]B.west);

\draw[-{Triangle[scale=1.5]}, line width=1] ([xshift=-1pt]B.west) -- (B.west);

\draw[-{Triangle[scale=1.5]}, line width=1] (B) -- (D);
\draw[-{Triangle[scale=1.5]}, line width=1] (A) -- (C);
\node[fill=white, draw=black, rounded corners=5pt] at (1.3*\dx, -0.4*\dy) {copy};
\node[fill=white, draw=black, rounded corners=5pt] at (0.65*\dx, 0*\dy) {inputs};
\node[fill=white, draw=black, rounded corners=5pt] at (1.89*\dx, 0*\dy) {output};

\end{tikzpicture}
    \caption{Schema of the sampling process of ciDATGAN.}
    \label{fig_ch3:sampling}
\end{figure}

The model needs to receive two inputs for the sampling phase: the Gaussian noise and the conditional inputs. The conditional inputs can either come from the feeder dataset $\bf T_F^{ci}$ or from a distributor dataset $\bf T_D^{ci}$. The generator will, thus, sample the complementary set of variables $\bf T^{c,synth}_{F\rightarrow D}$ based on the conditional inputs. Finally, it combines the conditional inputs and the synthetic data to deliver the final dataset. Figure~\ref{fig_ch3:sampling} shows a schematic representation of the sampling process.

\newpage
\section{Results}
\label{sec_ch3:results}

In this section, we investigate three questions:
\begin{itemize}
    \item Does ciDATGAN generate data well?, \emph{i.e.} are the generated data representative, and do they correspond to the original data?
    \item Can ciDATGAN correct bias?, \emph{i.e.} can the model use unbiased conditional inputs to correct the bias in the original data?
    \item Can ciDATGAN be used to generate large synthetic populations?, \emph{i.e.} can the model learn from a feeder dataset and efficiently generate a population using a large sample for the conditional inputs?
\end{itemize}

The first two questions are answered in Section~\ref{sec_ch3:cidatgan_vs_datgan}, the third question in Section~\ref{sec_ch3:large_pop}.

\subsection{ciDATGAN vs DATGAN}
\label{sec_ch3:cidatgan_vs_datgan}

This section answers the first two questions by comparing DATGAN and ciDATGAN. We first present the case study and the results assessments in the next section. Then, we answer the first two questions in order.

\subsubsection{Case study and results assessments}
\label{sec_ch3:case_study}

To evaluate the performance of ciDATGAN, we use a modified version of the LPMC dataset~\citep{hillel_recreating_2018} as the feeder dataset. A description of the variables included in this modified dataset is given in the appendix; see Table~\ref{tab_ch3:data_LPMC}. It consists of 18 variables and 16'904 trips. In the appendix, we also provide the DAG used in ciDATGAN for this dataset in Figure~\ref{fig_ch3:DAG_LPMC}. The conditional inputs are the variables \texttt{age}, \texttt{female}, and \texttt{hh\_region}.
We trained both ciDATGAN and DATGAN on the LPMC dataset five times. Then, we generated five different synthetic datasets for each of these models, thus giving 25 synthetic datasets per model. To assess the results, we use the same methodology as \cite{lederrey_datgan_2022}, namely: 
\begin{itemize}
    \item We compute frequency lists for each variable in the synthetic dataset and compare it to similar lists computed on the original dataset.
    \item We compute and compare frequency lists for each combination of two variables for the synthetic and original datasets.
    \item Finally, we train a LightGBM~\citep{ke_lightgbm_2017} model on all but one column and use the resulting model to predict the corresponding column in the original data. We do this for each column. We can then compare the results between the models trained on the synthetic datasets and those trained on the original dataset.
\end{itemize}

Details of this methodology are given by \cite{lederrey_datgan_2022}. For conciseness, in this chapter, we present only the results for the SRMSE~\citep{muller_hierarchical_2011}. This metric has been used thoroughly while investigating methods to generate synthetic populations. It consists in computing the RMSE on two frequency lists and dividing the value by the average value of the frequency list for the original data. Then, for each of the 25 datasets, we compute the average value for the given metric, and we can use boxplots to show the results over multiple generated datasets. 

\subsubsection{Data generation}
\label{sec_ch3:generate_data}

As shown by \cite{lederrey_datgan_2022}, DATGAN achieves state-of-the-art performance when generating synthetic tabular data regarding the generated data's quality and representativity. As such, if ciDATGAN can match or better DATGAN's performance, we demonstrate a positive answer to the first research question. To test this, we apply the results assessments on all the generated variables, \emph{i.e.} we discard the conditional inputs in the assessments since ciDATGAN uses the original values directly and would have an unfair advantage. 

\begin{figure}[!htb]
    \centering
    \begin{subfigure}{0.505\textwidth}
        \begin{tikzpicture}[inner frame sep=0]
    \begin{axis}[
        footnotesize,
        width=\textwidth,
        height=4.8cm,
        boxplot/draw direction=y,
        xtick={1,2},
        xticklabels={DATGAN, ciDATGAN},
        ylabel=SRMSE,
        every axis y label/.style={
            at={(ticklabel cs:0.5)},
            rotate=90,
            anchor=center,
            anchor=near ticklabel, 
            font=\footnotesize
        },
        yticklabel style={
                /pgf/number format/fixed,
                /pgf/number format/precision=2,
                rotate=90
        },
        ymajorgrids,
        axis line style={line width=0.1pt},
        ytick style={draw=none},
        xtick style={draw=none}
        ]
        \addplot+ [bplot style, fill=white, boxplot/every box/.style={fill=lightgray}] table [y index=0] {tikz/fig/obs/lvl1.dat}
        [above]
        node[text=black] at
        (boxplot box cs: \boxplotvalue{average}+0.005,0.5)
        {\footnotesize\pgfmathprintnumber{\boxplotvalue{average}}};
        \addplot+ [bplot style, fill=white, boxplot/every box/.style={fill=lightred}] table [y index=1] {tikz/fig/obs/lvl1.dat}
        [above]
        node[text=black] at
        (boxplot box cs: \boxplotvalue{average}+0.005,0.5)
        {\footnotesize\pgfmathprintnumber{\boxplotvalue{average}}};
        \end{axis}
\end{tikzpicture}
        \caption{Individual variables}
        \label{fig_ch3:lvl1}
    \end{subfigure}
    \hspace{-10pt}
    \begin{subfigure}{0.505\textwidth}
        \begin{tikzpicture}[inner frame sep=0]
        \begin{axis}[
        footnotesize,
        height=4.8cm,
        width=\textwidth,
        boxplot/draw direction=y,
        xtick={1,2},
        xticklabels={DATGAN, ciDATGAN},
        ylabel=SRMSE,
        every axis y label/.style={
            at={(ticklabel cs:0.5)},
            rotate=90,
            anchor=center,
            anchor=near ticklabel, 
            font=\footnotesize
        },
        yticklabel style={
                /pgf/number format/fixed,
                /pgf/number format/precision=2,
                rotate=90
        },
        ymajorgrids,
        axis line style={line width=0.1pt},
        ytick style={draw=none},
        xtick style={draw=none}
        ]
        \addplot+ [bplot style, fill=white, boxplot/every box/.style={fill=lightgray}] table [y index=0] {tikz/fig/obs/lvl2.dat}
        [above]
        node[text=black] at
        (boxplot box cs: \boxplotvalue{average}+0.005,0.5)
        {\footnotesize\pgfmathprintnumber{\boxplotvalue{average}}};
        \addplot+ [bplot style, fill=white, boxplot/every box/.style={fill=lightred}] table [y index=1] {tikz/fig/obs/lvl2.dat}
        [above]
        node[text=black] at
        (boxplot box cs: \boxplotvalue{average}+0.005,0.5)
        {\footnotesize\pgfmathprintnumber{\boxplotvalue{average}}};
        \end{axis}
\end{tikzpicture}
        \caption{Combinations of two variables}
        \label{fig_ch3:lvl2}
    \end{subfigure}
    \caption{Boxplots of SRMSE values for all the generated datasets from DATGAN and ciDATGAN. The white dot corresponds to the average value. Lower is better.}
    \label{fig_ch3:stats}
\end{figure}
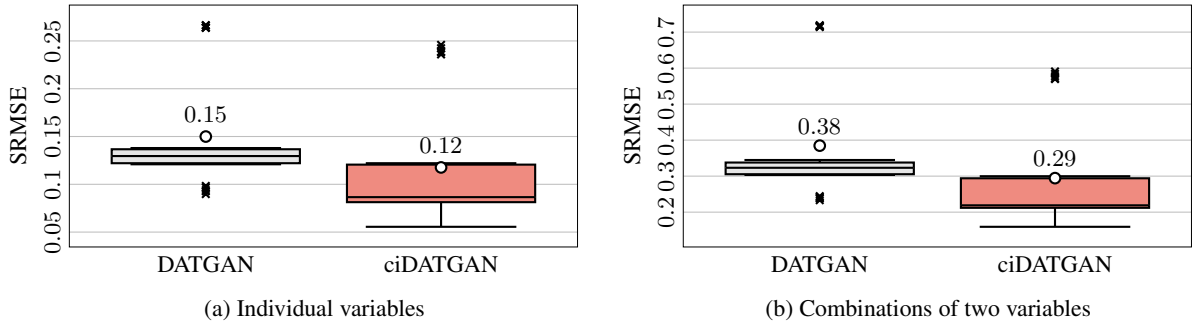

Results for the statistical assessments are given in Figure~\ref{fig_ch3:stats}. Figure~\ref{fig_ch3:lvl1} shows the results when comparing each variable independently, and Figure~\ref{fig_ch3:lvl2} shows the results when comparing each combination of two variables. We can see that, in both cases, ciDATGAN performs better than DATGAN. This difference comes from two different factors: \begin{inlinelist}
\item ciDATGAN has fewer variables to generate, thus, making it smaller and simpler to train; and 
\item ciDATGAN has direct access to true values from the datasets, thus, making the learning process easier. 
\end{inlinelist}

\begin{figure}[!htb]
    \centering
    \begin{subfigure}{0.505\textwidth}
        \begin{tikzpicture}[inner frame sep=0]
        \begin{axis}[
        footnotesize,
        height=4.8cm,
        width=\textwidth,
        boxplot/draw direction=y,
        ytick={0.08, 0.1, 0.12, 0.14},
        xtick={1,2},
        xticklabels={DATGAN, ciDATGAN},
        ylabel=Relative log loss,
        every axis y label/.style={
            at={(ticklabel cs:0.5)},
            rotate=90,
            anchor=center,
            anchor=near ticklabel, 
            font=\footnotesize
        },
        yticklabel style={
                /pgf/number format/fixed,
                /pgf/number format/precision=2,
                rotate=90
        },
        ymajorgrids,
        axis line style={line width=0.1pt},
        ytick style={draw=none},
        xtick style={draw=none}
        ]
        \addplot+ [bplot style, fill=white, boxplot/every box/.style={fill=lightgray}] table [y index=0] {tikz/fig/obs/ml_cat.dat}
        [above]
        node[text=black] at
        (boxplot box cs: \boxplotvalue{average}-0.013,0.5)
        {\footnotesize\pgfmathprintnumber[precision=3, fixed]{\boxplotvalue{average}}};
        
        \addplot+ [bplot style, fill=white, boxplot/every box/.style={fill=lightred}] table [y index=1] {tikz/fig/obs/ml_cat.dat}
        [above]
        node[text=black, style={/pgf/number format/precision=2}] at
        (boxplot box cs: \boxplotvalue{average}-0.013,0.5)
        {\footnotesize\pgfmathprintnumber[precision=3, fixed]{\boxplotvalue{average}}};
        \end{axis}
\end{tikzpicture}
        \caption{Categorical variables}
        \label{fig_ch3:ml_cat}
    \end{subfigure}
    \hspace{-10pt}
    \begin{subfigure}{0.505\textwidth}
        \begin{tikzpicture}[inner frame sep=0]
        \begin{axis}[
        footnotesize,
        height=4.8cm,
        width=\textwidth,
        boxplot/draw direction=y,
        ytick={0, 5, 10, 15},
        xtick={1,2},
        xticklabels={DATGAN, ciDATGAN},
        ylabel=Relative L2 error,
        every axis y label/.style={
            at={(ticklabel cs:0.5)},
            rotate=90,
            anchor=center,
            anchor=near ticklabel, 
            font=\footnotesize
        },
        yticklabel style={
                /pgf/number format/fixed,
                /pgf/number format/precision=2,
                rotate=90
        },
        ymajorgrids,
        axis line style={line width=0.1pt},
        ytick style={draw=none},
        xtick style={draw=none},
        ymax=18,
        ]
        \addplot+ [bplot style, fill=white, boxplot/every box/.style={fill=lightgray}] table [y index=0] {tikz/fig/obs/ml_cont.dat}
        [above]
        node[text=black] at
        (boxplot box cs: \boxplotvalue{average}+0.005,0.5)
        {\footnotesize\pgfmathprintnumber{\boxplotvalue{average}}};
        \addplot+ [bplot style, fill=white, boxplot/every box/.style={fill=lightred}] table [y index=1] {tikz/fig/obs/ml_cont.dat}
        [above]
        node[text=black] at
        (boxplot box cs: \boxplotvalue{average}+0.005,0.5)
        {\footnotesize\pgfmathprintnumber{\boxplotvalue{average}}};
        \end{axis}
\end{tikzpicture}
        \caption{Continuous variables}
        \label{fig_ch3:ml_cont}
    \end{subfigure}
    \caption{Boxplot of the ML efficacy metrics for all the generated datasets from DATGAN and ciDATGAN. The white dot corresponds to the average value. Lower is better.}
    \label{fig_ch3:ml}
\end{figure}
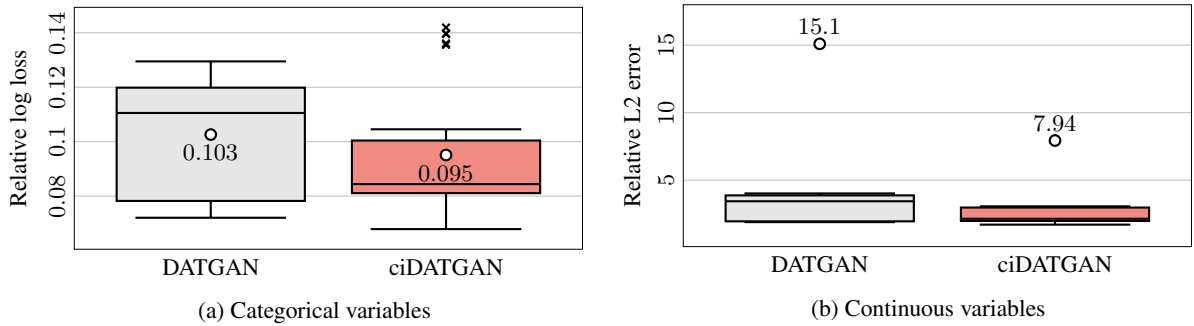

Results for the ML efficacy are given in Figure~\ref{fig_ch3:ml}. For categorical and continuous variables, ciDATGAN performs slightly better than DATGAN. This confirms the results obtained on the statistical metrics. Therefore, we can conclude that ciDATGAN better DATGAN's performances on similar datasets.

\subsubsection{Correcting bias}
\label{sec_ch3:correct_bias}

We have shown that ciDATGAN can generate data of equivalent or better quality to that generated by DATGAN. Thus, adding conditional inputs does not hinder the model's performance. This section investigates the use of ciDATGAN to remove bias from datasets using unbiased conditional inputs. Thus, we use the available LPMC dataset and bias it manually by removing: 
\begin{inlinelist}
    \item 70\% of males;
    \item 70\% of individuals 20 years old and younger; and
    \item 70\% of the households in the region "Central London".
\end{inlinelist}
This gives us a biased LPMC dataset of 8'437 individuals. The goal is to train both DATGAN and ciDATGAN on this biased dataset. Then, we generate data from these models such that the final dataset has the same size as the unbiased LPMC dataset. For ciDATGAN, we use the values from the unbiased LPMC dataset as the conditional inputs for the sampling phase. Finally, we test the generated data against the unbiased LPMC dataset. The assessments are done on a subset of the LPMC variables with higher correlations to the age, gender, and household region variables.

\begin{figure}[!htb]
    \begin{subfigure}{0.505\textwidth}
        \begin{tikzpicture}[inner frame sep=0]
    \begin{axis}[
        footnotesize,
        width=\textwidth,
        height=4.8cm,
        boxplot/draw direction=y,
        xtick={1,2,3},
        xticklabels={DATGAN, ciDATGAN, unbiased},
        xticklabel style={text width=2cm, align=center},
        ylabel=SRMSE,
        every axis y label/.style={
            at={(ticklabel cs:0.5)},
            rotate=90,
            anchor=center,
            anchor=near ticklabel, 
            font=\footnotesize
        },
        yticklabel style={
                /pgf/number format/fixed,
                /pgf/number format/precision=2,
                rotate=90
        },
        ymajorgrids,
        axis line style={line width=0.1pt},
        ytick style={draw=none},
        xtick style={draw=none}
        ]
        \addplot+ [bplot style, fill=white, boxplot/every box/.style={fill=lightgray}] table [y index=0] {tikz/fig/bias/lvl1.dat}
        [above]
        node[text=black] at
        (boxplot box cs: \boxplotvalue{average}-0.013,0.5)
        {\footnotesize\pgfmathprintnumber[precision=3, fixed]{\boxplotvalue{average}}};
        
        \addplot+ [bplot style, fill=white, boxplot/every box/.style={fill=lightred}] table [y index=1] {tikz/fig/bias/lvl1.dat}
        [above]
        node[text=black] at
        (boxplot box cs: \boxplotvalue{average}-0.0145,0.5)
        {\footnotesize\pgfmathprintnumber[precision=2, fixed]{\boxplotvalue{average}}};
        
        \addplot+ [bplot style, fill=white, boxplot/every box/.style={fill=lightgray}] table [y index=2] {tikz/fig/bias/lvl1.dat}
        [above]
        node[text=black] at
        (boxplot box cs: \boxplotvalue{average},0.5)
        {\footnotesize\pgfmathprintnumber[precision=2, fixed]{\boxplotvalue{average}}};
        
        \addplot+[mark=none, black, dashed, line width=1.5] coordinates {(0.6, 0.06711274120548238) (3.4, 0.06711274120548238)};
    \end{axis}
\end{tikzpicture}
        \caption{Individual variables}
        \label{fig_ch3:lvl1_bias}
    \end{subfigure}
    \hspace{-10pt}
    \begin{subfigure}{0.505\textwidth}
        \begin{tikzpicture}[inner frame sep=0]
        \begin{axis}[
        footnotesize,
        height=4.8cm,
        width=\textwidth,
        boxplot/draw direction=y,
        xtick={1,2,3},
        xticklabels={DATGAN, ciDATGAN, unbiased},
        xticklabel style={text width=2cm, align=center},
        ylabel=SRMSE,
        every axis y label/.style={
            at={(ticklabel cs:0.5)},
            rotate=90,
            anchor=center,
            anchor=near ticklabel, 
            font=\footnotesize
        },
        yticklabel style={
                /pgf/number format/fixed,
                /pgf/number format/precision=2,
                rotate=90
        },
        ymajorgrids,
        axis line style={line width=0.1pt},
        ytick style={draw=none},
        xtick style={draw=none}
        ]
        \addplot+ [bplot style, fill=white, boxplot/every box/.style={fill=lightgray}] table [y index=0] {tikz/fig/bias/lvl2.dat}
        [above]
        node[text=black] at
        (boxplot box cs: \boxplotvalue{average}-0.04,0.5)
        {\footnotesize\pgfmathprintnumber{\boxplotvalue{average}}};
        
        \addplot+ [bplot style, fill=white, boxplot/every box/.style={fill=lightred}] table [y index=1] {tikz/fig/bias/lvl2.dat}
        [above]
        node[text=black] at
        (boxplot box cs: \boxplotvalue{average}+0.005,0.5)
        {\footnotesize\pgfmathprintnumber{\boxplotvalue{average}}};
        
        \addplot+ [bplot style, fill=white, boxplot/every box/.style={fill=lightgray}] table [y index=2] {tikz/fig/bias/lvl2.dat}
        [above]
        node[text=black] at
        (boxplot box cs: \boxplotvalue{average},0.5)
        {\footnotesize\pgfmathprintnumber{\boxplotvalue{average}}};
        
        \addplot+[mark=none, black, dashed, line width=1.5] coordinates {(0.6, 0.15426885778705768) (3.4, 0.15426885778705768)};
        \end{axis}
\end{tikzpicture}
        \caption{Combinations of two variables}
        \label{fig_ch3:lvl2_bias}
    \end{subfigure}
    \caption{Boxplots of SRMSE values for all the generated datasets from DATGAN and ciDATGAN. The two models on the left have been trained on the biased LPMC dataset. The model on the right has been trained on the unbiased LPMC dataset. The black dashed line corresponds to the comparison between the biased and the unbiased LPMC datasets. The white dot corresponds to the average value. Lower is better.}
    \label{fig_ch3:stats_bias}
\end{figure}
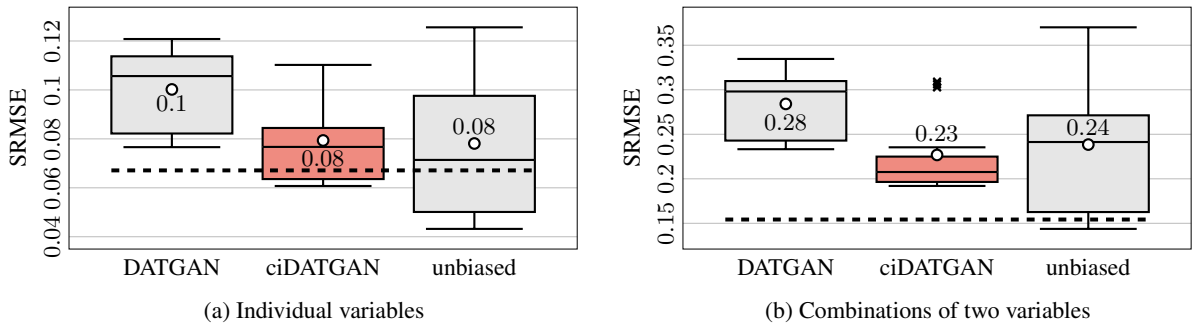

Results for the statistical assessments are given in Figure~\ref{fig_ch3:stats_bias}. On average, we see that the data generated by ciDATGAN is equivalent to that generated by DATGAN trained on the unbiased LPMC dataset. Both models outperform DATGAN trained on the biased LPMC dataset as the latter cannot correct for the bias introduced in the dataset.

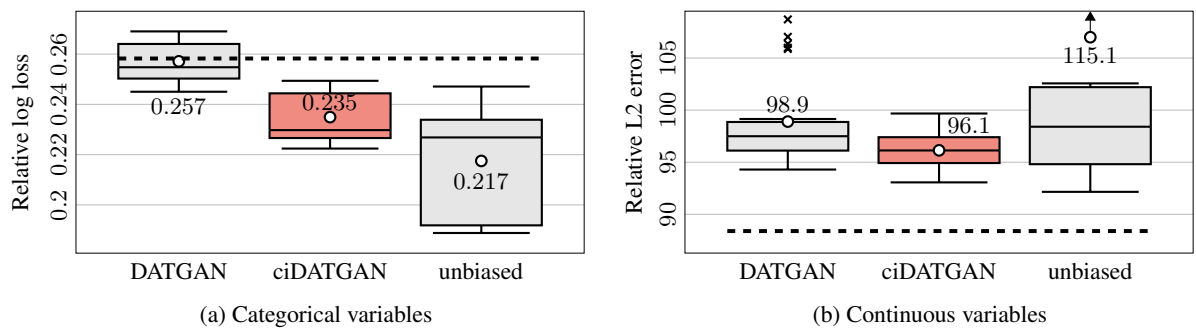
\begin{figure}[!htb]
    \begin{subfigure}{0.505\textwidth}
        \begin{tikzpicture}[inner frame sep=0]
        \begin{axis}[
        footnotesize,
        height=4.8cm,
        width=\textwidth,
        boxplot/draw direction=y,
        xtick={1,2,3},
        xticklabels={DATGAN, ciDATGAN, unbiased},
        xticklabel style={text width=2cm, align=center},
        ylabel=Relative log loss,
        every axis y label/.style={
            at={(ticklabel cs:0.5)},
            rotate=90,
            anchor=center,
            anchor=near ticklabel, 
            font=\footnotesize
        },
        yticklabel style={
                /pgf/number format/fixed,
                /pgf/number format/precision=2,
                rotate=90
        },
        ymajorgrids,
        axis line style={line width=0.1pt},
        ytick style={draw=none},
        xtick style={draw=none}
        ]
        \addplot+ [bplot style, fill=white, boxplot/every box/.style={fill=lightgray}] table [y index=0] {tikz/fig/bias/ml_cat.dat}
        [above]
        node[text=black] at
        (boxplot box cs: \boxplotvalue{average}-0.025,0.5)
        {\footnotesize\pgfmathprintnumber[precision=3, fixed]{\boxplotvalue{average}}};
        
        \addplot+ [bplot style, fill=white, boxplot/every box/.style={fill=lightred}] table [y index=1] {tikz/fig/bias/ml_cat.dat}
        [above]
        node[text=black, style={/pgf/number format/precision=2}] at
        (boxplot box cs: \boxplotvalue{average}-0.001,0.5)
        {\footnotesize\pgfmathprintnumber[precision=3, fixed]{\boxplotvalue{average}}};
        
        \addplot+ [bplot style, fill=white, boxplot/every box/.style={fill=lightgray}] table [y index=2] {tikz/fig/bias/ml_cat.dat}
        [above]
        node[text=black] at
        (boxplot box cs: \boxplotvalue{average}-0.015,0.5)
        {\footnotesize\pgfmathprintnumber[precision=3, fixed]{\boxplotvalue{average}}};
        
        \addplot+[mark=none, black, dashed, line width=1.5] coordinates {(0.6, 0.25825060463607596) (3.4, 0.25825060463607596)};        
        \end{axis}
\end{tikzpicture}
        \caption{Categorical variables}
        \label{fig_ch3:ml_cat_bias}
    \end{subfigure}
    \hspace{-10pt}
    \begin{subfigure}{0.505\textwidth}
        \begin{tikzpicture}[inner frame sep=0]
        \begin{axis}[
        footnotesize,
        height=4.8cm,
        width=\textwidth,
        ymax=109.5,
        boxplot/draw direction=y,
        xtick={1,2,3},
        xticklabels={DATGAN, ciDATGAN, unbiased},
        xticklabel style={text width=2cm, align=center},
        ylabel=Relative L2 error,
        every axis y label/.style={
            at={(ticklabel cs:0.5)},
            rotate=90,
            anchor=center,
            anchor=near ticklabel, 
            font=\footnotesize
        },
        yticklabel style={
                /pgf/number format/fixed,
                /pgf/number format/precision=2,
                rotate=90
        },
        ymajorgrids,
        axis line style={line width=0.1pt},
        ytick style={draw=none},
        xtick style={draw=none}
        ]
        \addplot+ [bplot style, fill=white, boxplot/every box/.style={fill=lightgray}] table [y index=0] {tikz/fig/bias/ml_cont.dat}
        [above]
        node[text=black] at
        (boxplot box cs: \boxplotvalue{average},0.5)
        {\footnotesize\pgfmathprintnumber[precision=1, fixed]{\boxplotvalue{average}}};
        
        \addplot+ [bplot style, fill=white, boxplot/every box/.style={fill=lightred}] table [y index=1] {tikz/fig/bias/ml_cont.dat}
        [above]
        node[text=black, style={/pgf/number format/precision=2}] at
        (boxplot box cs: \boxplotvalue{average}+0.7,0.75)
        {\footnotesize\pgfmathprintnumber[precision=1, fixed]{\boxplotvalue{average}}};
        
        \addplot+ [bplot style, fill=white, boxplot/every box/.style={fill=lightgray}] table [y index=2] {tikz/fig/bias/ml_cont.dat}
        [above]
        node[text=black] at
        (boxplot box cs: \boxplotvalue{average}-11.5,0.5)
        {\footnotesize\pgfmathprintnumber[precision=1, fixed]{\boxplotvalue{average}}};
        
        \addplot+[mark=none, black, dashed, line width=1.5] coordinates {(0.6, 88.3898986775722) (3.4, 88.3898986775722)}; 
        
        \addplot[mark=*, draw=black, fill=white, mark size=2pt, thick] coordinates {(3, 107)};
        \draw[-{Triangle}](axis cs:3, 107)--(axis cs:3, 109.5);
        \end{axis}
\end{tikzpicture}
        \caption{Continuous variables}
        \label{fig_ch3:ml_cont_bias}
    \end{subfigure}
    \caption{Boxplot of the ML efficacy metrics for all the generated datasets from DATGAN and ciDATGAN. The two models on the left have been trained on the biased LPMC dataset. The model on the right has been trained on the unbiased LPMC dataset. The black dashed line corresponds to the comparison between the biased and the unbiased LPMC datasets. The white dot corresponds to the average value. Lower is better.}
    \label{fig_ch3:ml_bias}
\end{figure}

Results for the ML efficacy are given in Figure~\ref{fig_ch3:ml_bias}. For the categorical variables, we observed that ciDATGAN and DATGAN trained on the unbiased LPMC dataset are performing well (Figure~\ref{fig_ch3:ml_cat_bias}). However, this is not the case for the continuous variables (Figure~\ref{fig_ch3:ml_cont_bias}). This result is explained because it is more difficult to generate representative continuous variables. 

\subsection{Population synthesis}
\label{sec_ch3:large_pop}

We need to compare the generated synthetic data with real data to answer the final question: \emph{Can ciDATGAN be used to generate large synthetic populations?} However, getting access to real micro-data is difficult due to privacy issues. Thus, we use aggregated data from the UK Census 2011, available on the nomis website\footnote{\url{https://www.nomisweb.co.uk/census/2011/data_finder}} as the ground truth. Nomis is a service provided by the Office for National Statistics (ONS), UK's largest independent producer of official statistics. The feeder dataset is a modified version of the LTDS dataset~\citep{hillel_recreating_2018} that only consists of variables for the individuals and their households. We selected a total of 10 (out of the 33) boroughs such that it gives a good representation of the London population. The feeder dataset, thus, consists of 8 variables and 29'158 individuals. The complete description of the dataset is given in Table~\ref{tab_ch3:data_LTDS} in the appendix. In addition, the DAG used for this dataset is provided in Figure~\ref{fig_ch3:DAG_LTDS}, in the appendix.

The distributor data has been created by drawing samples from the aggregate breakdowns of age and gender at the level of London boroughs from the nomis data. It contains a total of 2.7 million individuals for ten boroughs. Once the final synthetic dataset is generated, we aggregate the results and compare them against four overlapping variables, contained in both the LTDS and nomis control totals at the borough level: family composition, number of people per household, number of cars/vans per household, and ethnicity. 

Both the feeder and distributor datasets are defined on an individual level, \emph{i.e.} each row of the dataset corresponds to an individual. However, three of the four selected control totals are at the household level. Thus, directly comparing the synthetic population with the aggregated distributions is impossible. In the generated population, each individual is generated with a corresponding household size and several household level statistics, \emph{e.g.} number of vehicles in the household. To obtain unbiased aggregate household level statistics, we divide the household values by the number of household members. For example, to calculate the total number of vehicles in a borough, we would divide the number of vehicles in the household for each individual by the corresponding household size and sum over all individuals.

To validate the results, we compare ciDATGAN with DATGAN and an oversampled version of the LTDS datase. The latter is generated by oversampling the original LTDS dataset. For each selected borough, we oversample with replacement individuals from the same borough in the LTDS until the sample size is equal to the borough population, as recorded in the nomis data. We do not recommend doing this to augment the data since it introduces much redundancy. However, it will help understand if the synthetic data is on the same level of details and representativity as the original data in the feeder dataset. DATGAN is trained on the same feeder dataset as ciDATGAN. During the sampling phase, we randomly generate individuals until the sample size equals the borough population found in the nomis dataset. In the end, it gives us a total of three synthetic populations generated using three different methodologies such that each population reflects the nomis population for each borough. 

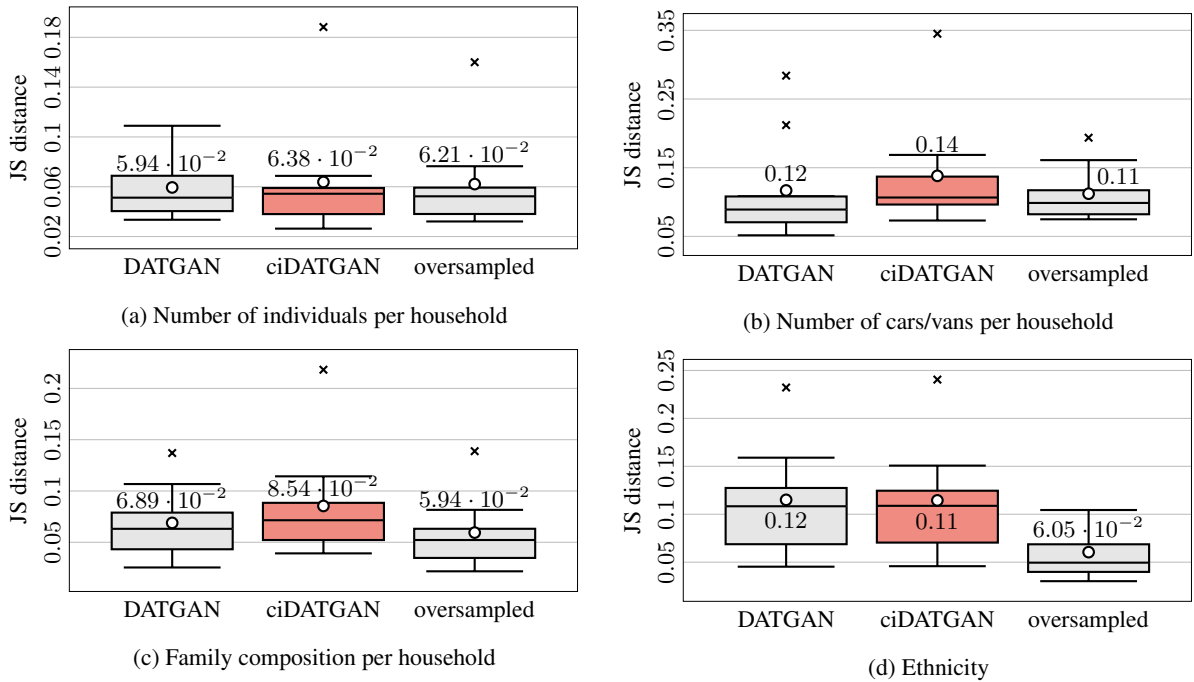
\begin{figure}[!htb]
    \centering
    \begin{subfigure}{0.505\textwidth}
        \begin{tikzpicture}[inner frame sep=0]
    \begin{axis}[
        footnotesize,
        width=\textwidth,
        height=4.8cm,
        boxplot/draw direction=y,
        xtick={1,2,3},
        xticklabels={DATGAN, ciDATGAN, oversampled},
        ytick={0.02, 0.06, 0.1, 0.14, 0.18},
        ylabel=JS distance,
        every axis y label/.style={
            at={(ticklabel cs:0.5)},
            rotate=90,
            anchor=center,
            anchor=near ticklabel, 
            font=\footnotesize
        },
        yticklabel style={
                /pgf/number format/fixed,
                /pgf/number format/precision=2,
                rotate=90
        },
        ymajorgrids,
        axis line style={line width=0.1pt},
        ytick style={draw=none},
        xtick style={draw=none}
        ]
        \addplot+ [bplot style, fill=white, boxplot/every box/.style={fill=lightgray}] table [y index=0] {tikz/fig/pop/hh_size.dat}
        [above]
        node[text=black] at
        (boxplot box cs: \boxplotvalue{average}+0.005,0.5)
        {\footnotesize\pgfmathprintnumber{\boxplotvalue{average}}};
        \addplot+ [bplot style, fill=white, boxplot/every box/.style={fill=lightred}] table [y index=1] {tikz/fig/pop/hh_size.dat}
        [above]
        node[text=black] at
        (boxplot box cs: \boxplotvalue{average}+0.005,0.5)
        {\footnotesize\pgfmathprintnumber{\boxplotvalue{average}}};
\addplot+ [bplot style, fill=white, boxplot/every box/.style={fill=lightgray}] table [y index=2] {tikz/fig/pop/hh_size.dat}
        [above]
        node[text=black] at
        (boxplot box cs: \boxplotvalue{average}+0.01,0.5)
        {\footnotesize\pgfmathprintnumber{\boxplotvalue{average}}};
        \end{axis}
\end{tikzpicture}
        \caption{Number of individuals per household}
        \label{fig_ch3:hh_size}
    \end{subfigure}
    \hspace{-10pt}
    \begin{subfigure}{0.505\textwidth}
        \begin{tikzpicture}[inner frame sep=0]
    \begin{axis}[
        footnotesize,
        width=\textwidth,
        height=4.8cm,
        boxplot/draw direction=y,
        xtick={1,2,3},
        xticklabels={DATGAN, ciDATGAN, oversampled},
        ytick={0.05, 0.15, 0.25, 0.35},
        ylabel=JS distance,
        every axis y label/.style={
            at={(ticklabel cs:0.5)},
            rotate=90,
            anchor=center,
            anchor=near ticklabel, 
            font=\footnotesize
        },
        yticklabel style={
                /pgf/number format/fixed,
                /pgf/number format/precision=2,
                rotate=90
        },
        ymajorgrids,
        axis line style={line width=0.1pt},
        ytick style={draw=none},
        xtick style={draw=none}
        ]
        \addplot+ [bplot style, fill=white, boxplot/every box/.style={fill=lightgray}] table [y index=0] {tikz/fig/pop/hh_car.dat}
        [above]
        node[text=black] at
        (boxplot box cs: \boxplotvalue{average},0.5)
        {\footnotesize\pgfmathprintnumber{\boxplotvalue{average}}};
        \addplot+ [bplot style, fill=white, boxplot/every box/.style={fill=lightred}] table [y index=1] {tikz/fig/pop/hh_car.dat}
        [above]
        node[text=black] at
        (boxplot box cs: \boxplotvalue{average}+0.02,0.5)
        {\footnotesize\pgfmathprintnumber{\boxplotvalue{average}}};
\addplot+ [bplot style, fill=white, boxplot/every box/.style={fill=lightgray}] table [y index=2] {tikz/fig/pop/hh_car.dat}
        [above]
        node[text=black] at
        (boxplot box cs: \boxplotvalue{average},0.75)
        {\footnotesize\pgfmathprintnumber{\boxplotvalue{average}}};
        \end{axis}
\end{tikzpicture}
        \caption{Number of cars/vans per household}
        \label{fig_ch3:hh_car}
    \end{subfigure}
    \\
    \begin{subfigure}{0.505\textwidth}
        \begin{tikzpicture}[inner frame sep=0]
    \begin{axis}[
        footnotesize,
        width=\textwidth,
        height=4.8cm,
        boxplot/draw direction=y,
        xtick={1,2,3},
        xticklabels={DATGAN, ciDATGAN, oversampled},
        ylabel=JS distance,
        every axis y label/.style={
            at={(ticklabel cs:0.5)},
            rotate=90,
            anchor=center,
            anchor=near ticklabel, 
            font=\footnotesize
        },
        yticklabel style={
                /pgf/number format/fixed,
                /pgf/number format/precision=2,
                rotate=90
        },
        ymajorgrids,
        axis line style={line width=0.1pt},
        ytick style={draw=none},
        xtick style={draw=none}
        ]
        \addplot+ [bplot style, fill=white, boxplot/every box/.style={fill=lightgray}] table [y index=0] {tikz/fig/pop/hh_comp.dat}
        [above]
        node[text=black] at
        (boxplot box cs: \boxplotvalue{average}+0.005,0.5)
        {\footnotesize\pgfmathprintnumber{\boxplotvalue{average}}};
        \addplot+ [bplot style, fill=white, boxplot/every box/.style={fill=lightred}] table [y index=1] {tikz/fig/pop/hh_comp.dat}
        [above]
        node[text=black] at
        (boxplot box cs: \boxplotvalue{average},0.5)
        {\footnotesize\pgfmathprintnumber{\boxplotvalue{average}}};
\addplot+ [bplot style, fill=white, boxplot/every box/.style={fill=lightgray}] table [y index=2] {tikz/fig/pop/hh_comp.dat}
        [above]
        node[text=black] at
        (boxplot box cs: \boxplotvalue{average}+0.015,0.5)
        {\footnotesize\pgfmathprintnumber{\boxplotvalue{average}}};
        \end{axis}
\end{tikzpicture}
        \caption{Family composition per household}
        \label{fig_ch3:hh_comp}
    \end{subfigure}
    \hspace{-10pt}
    \begin{subfigure}{0.505\textwidth}
        \begin{tikzpicture}[inner frame sep=0]
    \begin{axis}[
        footnotesize,
        width=\textwidth,
        height=4.8cm,
        boxplot/draw direction=y,
        xtick={1,2,3},
        xticklabels={DATGAN, ciDATGAN, oversampled},
        ylabel=JS distance,
        every axis y label/.style={
            at={(ticklabel cs:0.5)},
            rotate=90,
            anchor=center,
            anchor=near ticklabel, 
            font=\footnotesize
        },
        yticklabel style={
                /pgf/number format/fixed,
                /pgf/number format/precision=2,
                rotate=90
        },
        ymajorgrids,
        axis line style={line width=0.1pt},
        ytick style={draw=none},
        xtick style={draw=none}
        ]
        \addplot+ [bplot style, fill=white, boxplot/every box/.style={fill=lightgray}] table [y index=0] {tikz/fig/pop/ethn.dat}
        [above]
        node[text=black] at
        (boxplot box cs: \boxplotvalue{average}-0.04,0.5)
        {\footnotesize\pgfmathprintnumber{\boxplotvalue{average}}};
        \addplot+ [bplot style, fill=white, boxplot/every box/.style={fill=lightred}] table [y index=1] {tikz/fig/pop/ethn.dat}
        [above]
        node[text=black] at
        (boxplot box cs: \boxplotvalue{average}-0.04,0.5)
        {\footnotesize\pgfmathprintnumber{\boxplotvalue{average}}};
\addplot+ [bplot style, fill=white, boxplot/every box/.style={fill=lightgray}] table [y index=2] {tikz/fig/pop/ethn.dat}
        [above]
        node[text=black] at
        (boxplot box cs: \boxplotvalue{average}+0.005,0.5)
        {\footnotesize\pgfmathprintnumber{\boxplotvalue{average}}};
        \end{axis}
\end{tikzpicture}
        \caption{Ethnicity}
        \label{fig_ch3:ethn}
    \end{subfigure}
    \caption{Boxplot of the JS distance comparing the three synthetic data methodologies against the nomis data on the four selected variables. The white dot corresponds to the average value. Lower is better.}
    \label{fig_ch3:results_pop}
\end{figure}

We proposed systematic methods based on statistical assessments and ML efficacy methods to compare synthetic data in the previous chapter. However, we compare generated tabular data with aggregate data in this section. Thus, it is not possible to use the same assessment methods. However, all the overlapping variables are categorical, see Table~\ref{tab_ch3:data_LTDS}. Thus, the aggregate distributions are all discrete, and we can use the Jensen-Shannon (JS) distance~\citep{lin_divergence_1991} to measure the distance between these aggregate distributions. We define the JS distance, the square root of the JS divergence, between two probability arrays $P$ and $Q$ by:
\begin{equation}
JS(P,Q) = \sqrt{\frac{KL(P,M) + KL(Q,M)}{2}}
\end{equation}
where $M$ is the pointwise mean of $P$ and $Q$, and $KL(\cdot,\cdot)$ is the KL divergence defined by
\begin{equation}
KL(P,Q) = \left(P, Q\right) = \sum_{x\in\mathcal{X}} P(x) \log\left(\frac{P(x)}{Q(x)}\right). 
\end{equation}
The results are given in Figure~\ref{fig_ch3:results_pop}. It shows that the three methodologies generate roughly equivalent synthetic populations compared to the aggregate control totals in the nomis data. Thus, it shows that the addition of the conditional inputs in ciDATGAN does not hinder its performance compared to DATGAN. However, it is interesting to note that the oversampled LTDS dataset generates slightly more representative data for the family composition and ethnicity. 

This behavior is not surprising due to the collection process of the LTDS dataset. Indeed, it is intended to use as representative a sample as possible. Thus, LTDS is a good sample of the true population, as it is given in the nomis data. In addition, since the feeder dataset does not contain any bias, the distributor dataset follows the same distributions. Thus, it does not improve the final synthetic population generated by ciDATGAN. We show this similarity between the aggregated nomis data and the LTDS data in Figures~\ref{fig_ch3:ethn_check} and \ref{fig_ch3:cars_check}. 

Figure~\ref{fig_ch3:ethn_check} shows the ethnicity distribution for these two datasets on an aggregated level for each selected borough. The same trends are found in both datasets. For example, we can see that Brent has fewer White people and more Asian people than the other boroughs in both datasets. Figure~\ref{fig_ch3:cars_check} shows similar distributions for the number of cars/vans per household. We see that, in both datasets, the boroughs close to the center of London (Camden and Westminster) have fewer cars in their household than the others. However, we can see that the LTDS data tends to have more households with one vehicle and fewer without vehicles. This will skew the results of the synthetic data since the models tend to generate more households with one vehicle instead of zero. However, since we are comparing the generated data against the LTDS data, both data should suffer equally from this. However, DATGAN has already shown its capabilities of generating representative synthetic data. Thus, we want to investigate how the conditionality of ciDATGAN can improve the generation process when the model is dealing with poorly designed data.

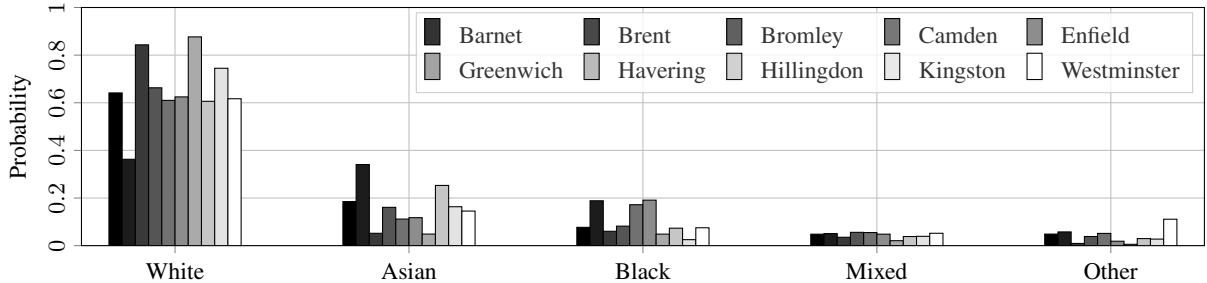
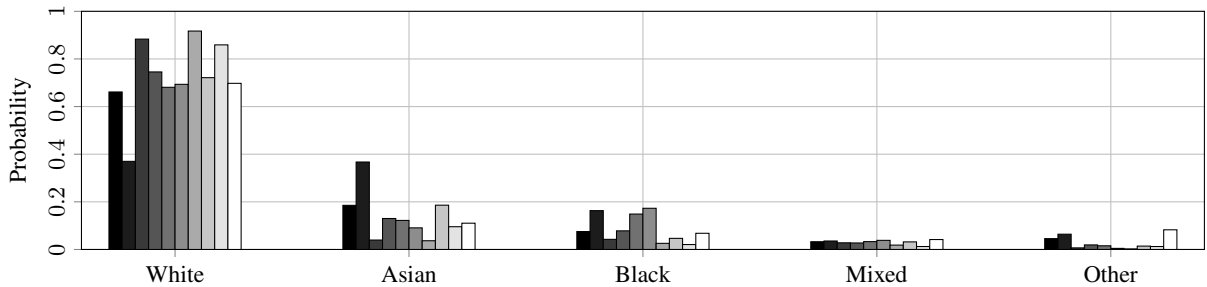
\begin{figure}[!htb]
    \centering
    \begin{subfigure}{\textwidth}
        \begin{tikzpicture}[inner frame sep=0]

    \begin{axis}[
        footnotesize,
        width=\textwidth,
        height=4.75cm,
        xtick={0,1,2,3,4},
        xticklabels={White, Asian, Black, Mixed, Other},
        ytick={0, 0.2, 0.4, 0.6,  0.8, 1},
        ylabel=Probability,
        every axis y label/.style={
            at={(ticklabel cs:0.5)},
            rotate=90,
            anchor=center,
            anchor=near ticklabel, 
            font=\footnotesize
        },
        yticklabel style={
                /pgf/number format/fixed,
                /pgf/number format/precision=2,
                rotate=90
        },
        ybar=0,
        bar width=5pt,
        legend columns=5,
        legend cell align={left},
        legend style={
            fill opacity=0.8, 
            draw opacity=1, 
            text opacity=1, 
            draw=white!80!black, 
            font=\footnotesize,
            /tikz/every even column/.append style={column sep=0.2cm},
            at={(0.99,0.98)},
            anchor=north east
        },
        tick align=outside,
        tick pos=left,
        xmajorgrids,
        ymajorgrids,
        legend image code/.code={\draw [#1] (0cm,-0.1cm) rectangle (0.2cm,0.25cm); },
        ymin=0,
        ymax=1
    ]
    
    \foreach [evaluate=\i as \n using (\i-1)*100/(9)] \i in {1,2,...,10} {%
    	\edef\temp{\noexpand\addplot[draw=black,fill=white!\n!black] table[x index=0,y index=\i]{tikz/fig/nomis/ethn_nomis.dat};
    	}
    	\temp;
    }

    \legend{Barnet, Brent, Bromley, Camden, Enfield, Greenwich, Havering, Hillingdon, Kingston, Westminster}
    
    \end{axis}

\end{tikzpicture}
        \caption{Nomis data}
        \label{fig_ch3:ethn_nomis}
    \end{subfigure}
    \\
    \begin{subfigure}{\textwidth}
        \begin{tikzpicture}[inner frame sep=0]

    \begin{axis}[
        footnotesize,
        width=\textwidth,
        height=4.75cm,
        xtick={0,1,2,3,4},
        xticklabels={White, Asian, Black, Mixed, Other},
        ytick={0, 0.2, 0.4, 0.6,  0.8, 1},
        ylabel=Probability,
        every axis y label/.style={
            at={(ticklabel cs:0.5)},
            rotate=90,
            anchor=center,
            anchor=near ticklabel, 
            font=\footnotesize
        },
        yticklabel style={
                /pgf/number format/fixed,
                /pgf/number format/precision=2,
                rotate=90
        },
        ybar=0,
        bar width=5pt,
        legend columns=5,
        legend cell align={left},
        legend style={
            fill opacity=0.8, 
            draw opacity=1, 
            text opacity=1, 
            draw=white!80!black, 
            font=\footnotesize,
            /tikz/every even column/.append style={column sep=0.2cm},
            at={(0.99,0.98)},
            anchor=north east
        },
        tick align=outside,
        tick pos=left,
        xmajorgrids,
        ymajorgrids,
        legend image code/.code={\draw [#1] (0cm,-0.1cm) rectangle (0.2cm,0.25cm); },
        ymin=0,
        ymax=1
    ]
    
    \foreach [evaluate=\i as \n using (\i-1)*100/(9)] \i in {1,2,...,10} {%
    	\edef\temp{\noexpand\addplot[draw=black,fill=white!\n!black] table[x index=0,y index=\i]{tikz/fig/nomis/ethn_ltds.dat};
    	}
    	\temp;
    }

    
    \end{axis}

\end{tikzpicture}
        \caption{LTDS data}
        \label{fig_ch3:ethn_ltds}
    \end{subfigure}
    \caption{Distributions of the individuals' ethnicity for each borough.}
    \label{fig_ch3:ethn_check}
\end{figure}
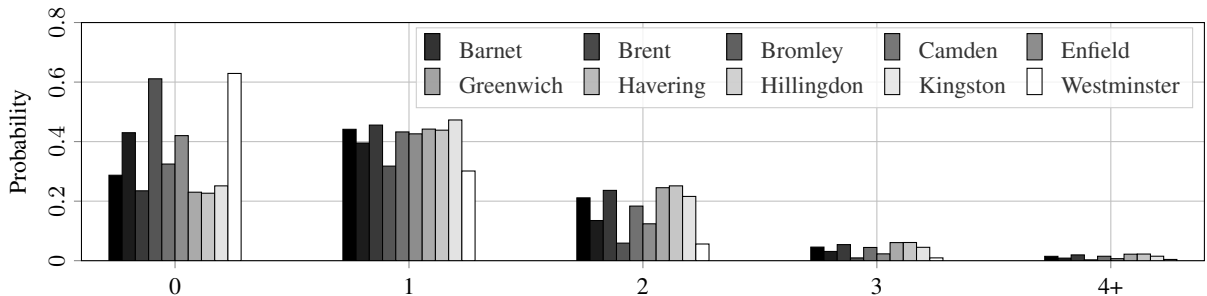
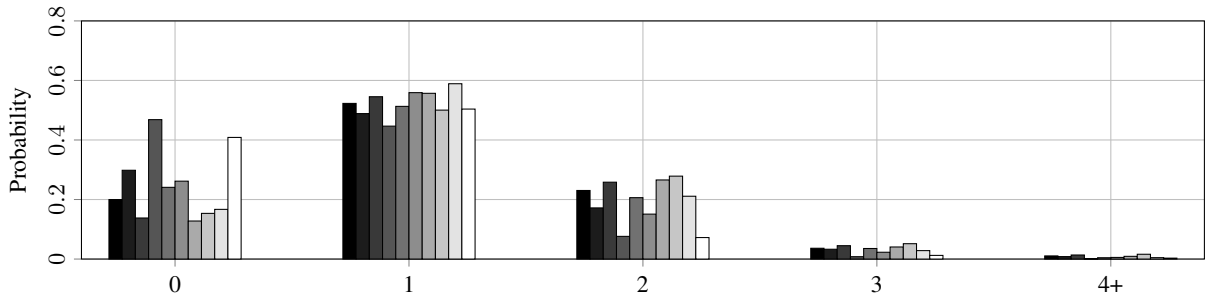
\begin{figure}[!htb]
    \centering
    \begin{subfigure}{\textwidth}
        \begin{tikzpicture}[inner frame sep=0]

    \begin{axis}[
        footnotesize,
        width=\textwidth,
        height=4.75cm,
        xtick={0,1,2,3,4},
        xticklabels={0,1,2,3,4+},
        ytick={0, 0.2, 0.4, 0.6, 0.8},
        ylabel=Probability,
        every axis y label/.style={
            at={(ticklabel cs:0.5)},
            rotate=90,
            anchor=center,
            anchor=near ticklabel, 
            font=\footnotesize
        },
        yticklabel style={
                /pgf/number format/fixed,
                /pgf/number format/precision=2,
                rotate=90
        },
        ybar=0,
        bar width=5pt,
        legend columns=5,
        legend cell align={left},
        legend style={
            fill opacity=0.8, 
            draw opacity=1, 
            text opacity=1, 
            draw=white!80!black, 
            font=\footnotesize,
            /tikz/every even column/.append style={column sep=0.2cm},
            at={(0.99,0.98)},
            anchor=north east
        },
        tick align=outside,
        tick pos=left,
        xmajorgrids,
        ymajorgrids,
        legend image code/.code={\draw [#1] (0cm,-0.1cm) rectangle (0.2cm,0.25cm); },
        ymin=0,
        ymax=0.8
    ]
    
    \foreach [evaluate=\i as \n using (\i-1)*100/(9)] \i in {1,2,...,10} {%
    	\edef\temp{\noexpand\addplot[draw=black,fill=white!\n!black] table[x index=0,y index=\i]{tikz/fig/nomis/car_nomis.dat};
    	}
    	\temp;
    }

    \legend{Barnet, Brent, Bromley, Camden, Enfield, Greenwich, Havering, Hillingdon, Kingston, Westminster}
    
    \end{axis}

\end{tikzpicture}
        \caption{Nomis data}
        \label{fig_ch3:car_nomis}
    \end{subfigure}
    \\
    \begin{subfigure}{\textwidth}
        \begin{tikzpicture}[inner frame sep=0]

    \begin{axis}[
        footnotesize,
        width=\textwidth,
        height=4.75cm,
        xtick={0,1,2,3,4},
        xticklabels={0, 1, 2, 3, 4+},
        ytick={0, 0.2, 0.4, 0.6, 0.8},
        ylabel=Probability,
        every axis y label/.style={
            at={(ticklabel cs:0.5)},
            rotate=90,
            anchor=center,
            anchor=near ticklabel, 
            font=\footnotesize
        },
        yticklabel style={
                /pgf/number format/fixed,
                /pgf/number format/precision=2,
                rotate=90
        },
        ybar=0,
        bar width=5pt,
        legend columns=4,
        legend cell align={left},
        legend style={
            fill opacity=0.8, 
            draw opacity=1, 
            text opacity=1, 
            draw=white!80!black, 
            font=\footnotesize,
            /tikz/every even column/.append style={column sep=0.2cm},
            at={(0.99,0.98)},
            anchor=north east
        },
        tick align=outside,
        tick pos=left,
        xmajorgrids,
        ymajorgrids,
        legend image code/.code={\draw [#1] (0cm,-0.1cm) rectangle (0.2cm,0.25cm); },
        ymin=0,
        ymax=0.8
    ]
    
    \foreach [evaluate=\i as \n using (\i-1)*100/(9)] \i in {1,2,...,10} {%
    	\edef\temp{\noexpand\addplot[draw=black,fill=white!\n!black] table[x index=0,y index=\i]{tikz/fig/nomis/car_ltds.dat};
    	}
    	\temp;
    }

    
    \end{axis}

\end{tikzpicture}
        \caption{LTDS data}
        \label{fig_ch3:car_ltds}
    \end{subfigure}
    \caption{Distributions of the number of cars/vans per household for each borough.}
    \label{fig_ch3:cars_check}
\end{figure}

\clearpage

\begin{figure}[!htb]
    \centering
    \begin{subfigure}{0.505\textwidth}
        \begin{tikzpicture}[inner frame sep=0]
    \begin{axis}[
        footnotesize,
        width=\textwidth,
        height=4.8cm,
        boxplot/draw direction=y,
        xtick={1,2,3},
        xticklabels={DATGAN, ciDATGAN, oversampled},
        ylabel=JS distance,
        every axis y label/.style={
            at={(ticklabel cs:0.5)},
            rotate=90,
            anchor=center,
            anchor=near ticklabel, 
            font=\footnotesize
        },
        yticklabel style={
                /pgf/number format/fixed,
                /pgf/number format/precision=2,
                rotate=90
        },
        ymajorgrids,
        axis line style={line width=0.1pt},
        ytick style={draw=none},
        xtick style={draw=none}
        ]
        \addplot+ [bplot style, fill=white, boxplot/every box/.style={fill=lightgray}] table [y index=0] {tikz/fig/pop_del/hh_size.dat}
        [above]
        node[text=black] at
        (boxplot box cs: \boxplotvalue{average}-0.04,0.5)
        {\footnotesize\pgfmathprintnumber{\boxplotvalue{average}}};
        \addplot+ [bplot style, fill=white, boxplot/every box/.style={fill=lightred}] table [y index=1] {tikz/fig/pop_del/hh_size.dat}
        [above]
        node[text=black] at
        (boxplot box cs: \boxplotvalue{average}+0.015,0.5)
        {\footnotesize\pgfmathprintnumber{\boxplotvalue{average}}};
\addplot+ [bplot style, fill=white, boxplot/every box/.style={fill=lightgray}] table [y index=2] {tikz/fig/pop_del/hh_size.dat}
        [above]
        node[text=black] at
        (boxplot box cs: \boxplotvalue{average}-0.04,0.5)
        {\footnotesize\pgfmathprintnumber{\boxplotvalue{average}}};
        \end{axis}
\end{tikzpicture}
        \caption{Number of individuals per household}
        \label{fig_ch3:hh_size_del}
    \end{subfigure}
    \hspace{-10pt}
    \begin{subfigure}{0.505\textwidth}
        \begin{tikzpicture}[inner frame sep=0]
    \begin{axis}[
        footnotesize,
        width=\textwidth,
        height=4.8cm,
        boxplot/draw direction=y,
        xtick={1,2,3},
        xticklabels={DATGAN, ciDATGAN, oversampled},
        ymax=0.31,
        ymin=0.03,
        ylabel=JS distance,
        every axis y label/.style={
            at={(ticklabel cs:0.5)},
            rotate=90,
            anchor=center,
            anchor=near ticklabel, 
            font=\footnotesize
        },
        yticklabel style={
                /pgf/number format/fixed,
                /pgf/number format/precision=2,
                rotate=90
        },
        ymajorgrids,
        axis line style={line width=0.1pt},
        ytick style={draw=none},
        xtick style={draw=none}
        ]
        \addplot+ [bplot style, fill=white, boxplot/every box/.style={fill=lightgray}] table [y index=0] {tikz/fig/pop_del/hh_car.dat}
        [above]
        node[text=black] at
        (boxplot box cs: \boxplotvalue{average}-0.055,0.5)
        {\footnotesize\pgfmathprintnumber{\boxplotvalue{average}}};
        \addplot+ [bplot style, fill=white, boxplot/every box/.style={fill=lightred}] table [y index=1] {tikz/fig/pop_del/hh_car.dat}
        [above]
        node[text=black] at
        (boxplot box cs: \boxplotvalue{average}+0.025,0.5)
        {\footnotesize\pgfmathprintnumber{\boxplotvalue{average}}};
\addplot+ [bplot style, fill=white, boxplot/every box/.style={fill=lightgray}] table [y index=2] {tikz/fig/pop_del/hh_car.dat}
        [above]
        node[text=black] at
        (boxplot box cs: \boxplotvalue{average}+0.005,0.72)
        {\footnotesize\pgfmathprintnumber{\boxplotvalue{average}}};
        \end{axis}
\end{tikzpicture}
        \caption{Number of car/van per household}
        \label{fig_ch3:hh_car_del}
    \end{subfigure}
    \\
    \begin{subfigure}{0.505\textwidth}
        \begin{tikzpicture}[inner frame sep=0]
    \begin{axis}[
        footnotesize,
        width=\textwidth,
        height=4.8cm,
        boxplot/draw direction=y,
        xtick={1,2,3},
        xticklabels={DATGAN, ciDATGAN, oversampled},
        ylabel=JS distance,
        every axis y label/.style={
            at={(ticklabel cs:0.5)},
            rotate=90,
            anchor=center,
            anchor=near ticklabel, 
            font=\footnotesize
        },
        yticklabel style={
                /pgf/number format/fixed,
                /pgf/number format/precision=2,
                rotate=90
        },
        ymajorgrids,
        axis line style={line width=0.1pt},
        ytick style={draw=none},
        xtick style={draw=none}
        ]
        \addplot+ [bplot style, fill=white, boxplot/every box/.style={fill=lightgray}] table [y index=0] {tikz/fig/pop_del/hh_comp.dat}
        [above]
        node[text=black] at
        (boxplot box cs: \boxplotvalue{average}+0.005,0.5)
        {\footnotesize\pgfmathprintnumber{\boxplotvalue{average}}};
        \addplot+ [bplot style, fill=white, boxplot/every box/.style={fill=lightred}] table [y index=1] {tikz/fig/pop_del/hh_comp.dat}
        [above]
        node[text=black] at
        (boxplot box cs: \boxplotvalue{average}+0.005,0.5)
        {\footnotesize\pgfmathprintnumber{\boxplotvalue{average}}};
\addplot+ [bplot style, fill=white, boxplot/every box/.style={fill=lightgray}] table [y index=2] {tikz/fig/pop_del/hh_comp.dat}
        [above]
        node[text=black] at
        (boxplot box cs: \boxplotvalue{average}+0.005,0.5)
        {\footnotesize\pgfmathprintnumber{\boxplotvalue{average}}};
        \end{axis}
\end{tikzpicture}
        \caption{Family composition per household}
        \label{fig_ch3:hh_comp_del}
    \end{subfigure}
    \hspace{-10pt}
    \begin{subfigure}{0.505\textwidth}
        \begin{tikzpicture}[inner frame sep=0]
    \begin{axis}[
        footnotesize,
        width=\textwidth,
        height=4.8cm,
        boxplot/draw direction=y,
        xtick={1,2,3},
        xticklabels={DATGAN, ciDATGAN, oversampled},
        ylabel=JS distance,
        every axis y label/.style={
            at={(ticklabel cs:0.5)},
            rotate=90,
            anchor=center,
            anchor=near ticklabel, 
            font=\footnotesize
        },
        yticklabel style={
                /pgf/number format/fixed,
                /pgf/number format/precision=2,
                rotate=90
        },
        ymajorgrids,
        axis line style={line width=0.1pt},
        ytick style={draw=none},
        xtick style={draw=none}
        ]
        \addplot+ [bplot style, fill=white, boxplot/every box/.style={fill=lightgray}] table [y index=0] {tikz/fig/pop_del/ethn.dat}
        [above]
        node[text=black] at
        (boxplot box cs: \boxplotvalue{average}-0.04,0.5)
        {\footnotesize\pgfmathprintnumber{\boxplotvalue{average}}};
        \addplot+ [bplot style, fill=white, boxplot/every box/.style={fill=lightred}] table [y index=1] {tikz/fig/pop_del/ethn.dat}
        [above]
        node[text=black] at
        (boxplot box cs: \boxplotvalue{average}-0.037,0.5)
        {\footnotesize\pgfmathprintnumber{\boxplotvalue{average}}};
\addplot+ [bplot style, fill=white, boxplot/every box/.style={fill=lightgray}] table [y index=2] {tikz/fig/pop_del/ethn.dat}
        [above]
        node[text=black] at
        (boxplot box cs: \boxplotvalue{average}+0.045,0.5)
        {\footnotesize\pgfmathprintnumber{\boxplotvalue{average}}};
        \end{axis}
\end{tikzpicture}
        \caption{Ethnicity}
        \label{fig_ch3:ethn_del}
    \end{subfigure}
    \caption{Boxplot of the JS distance comparing the three synthetic data methodologies, using the biased LTDS dataset, against the nomis data on the four selected variables. The white dot corresponds to the average value. Lower is better.}
    \label{fig_ch3:results_del_pop}
\end{figure}
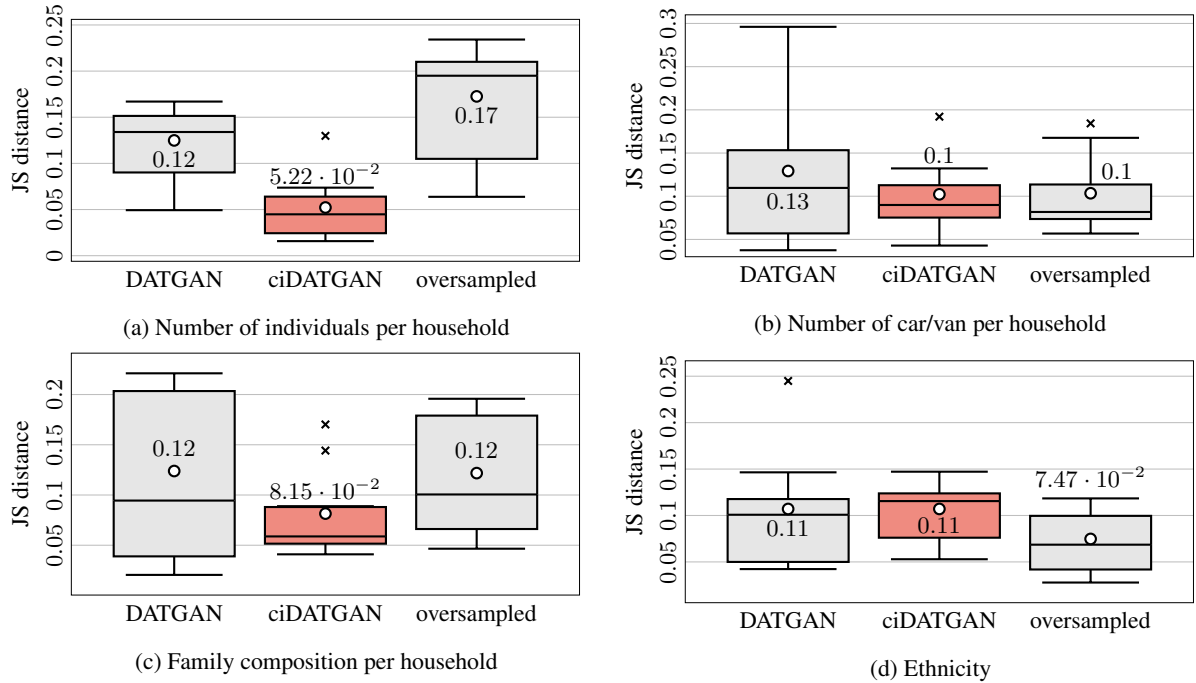

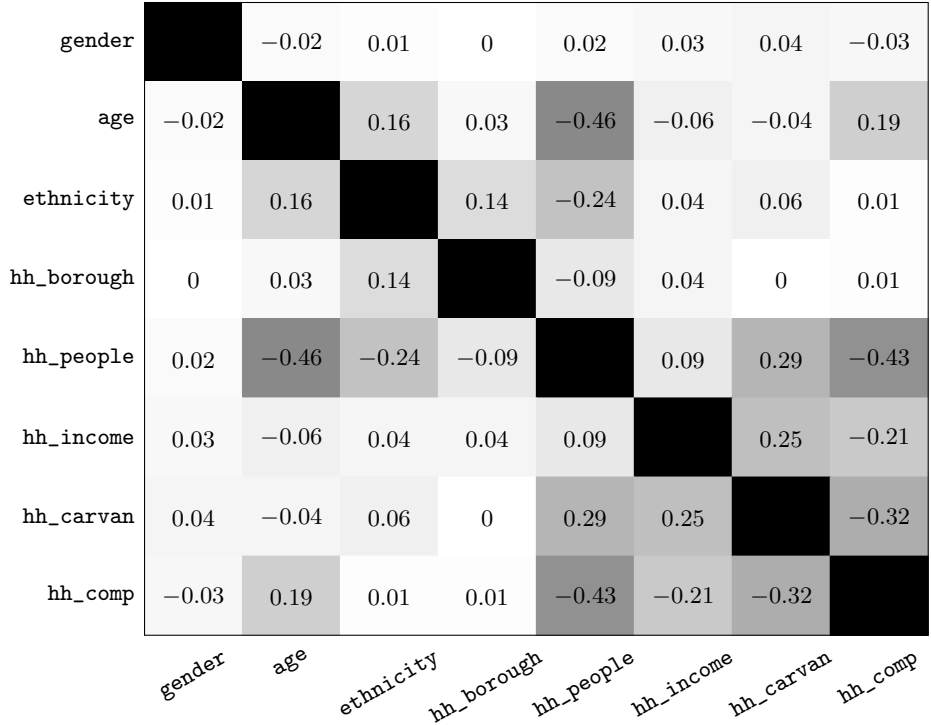
\begin{figure}[!htb]
\centering
\begin{tikzpicture}
    \begin{axis}[
        view={0}{90},   
        colormap={blackwhite}{gray(0cm)=(0); gray(0.5cm)=(1); gray(1cm)=(0)},
        %
        enlargelimits=false,
        axis on top,
        point meta min=-1,
        point meta max=1,
        xtick={0,1,2,3,4,5,6,7},
        xticklabels={\texttt{gender},\texttt{age},\texttt{ethnicity},\texttt{hh\_borough},\texttt{hh\_people},\texttt{hh\_income},\texttt{hh\_carvan},\texttt{hh\_comp}},
        ytick={0,1,2,3,4,5,6,7},
        yticklabels={\texttt{gender},\texttt{age},\texttt{ethnicity},\texttt{hh\_borough},\texttt{hh\_people},\texttt{hh\_income},\texttt{hh\_carvan},\texttt{hh\_comp}},
        xticklabel style={
            rotate=30,
            font=\footnotesize
        },
        yticklabel style={
            font=\footnotesize
        },
        ytick style={draw=none},
        xtick style={draw=none},
        y dir=reverse,
        nodes near coords={\pgfmathprintnumber[precision=2, fixed]\pgfplotspointmeta},
        nodes near coords align = south,
        every node near coord/.append style={xshift=0pt,yshift=-7pt, font=\footnotesize},
        height=10cm,
        width=12cm,
        %
    ]


        \addplot [matrix plot*,point meta=explicit] file {tikz/fig/nomis/corr.dat};
    \end{axis}
\end{tikzpicture}
\caption{Pearson's correlation matrix of the LTDS dataset.}
\label{fig_ch3:corr_mat}
\end{figure}

In Section~\ref{sec_ch3:correct_bias}, we have shown that ciDATGAN outperforms DATGAN when correcting for bias. Thus, we investigate if ciDATGAN provides similar results on a larger synthetic population. First, we need to bias the LTDS dataset. However, since we compare the synthetic population aggregated at the borough level, we do not bias the sample by boroughs alone. In addition, gender has a low correlation with the other variables, as shown in Figure~\ref{fig_ch3:corr_mat}. Thus, we bias the dataset on age. For each borough, we randomly select one of the three age categories: young (below 25 y.o.), middle (between 25 and 55 y.o.), and old (above 55 y.o.). We remove 95\% of the individuals from the other age categories. The final biased LTDS dataset contains a total of 10'009 individuals. Similar to the previous experiment, we train DATGAN and ciDATGAN on this biased dataset without changing the DAG in Figure~\ref{fig_ch3:DAG_LTDS} in the appendix. ciDATGAN is generating the new synthetic population using the previously defined distributor dataset, while the other two methodologies use the same sampling process. Results are given in Figure~\ref{fig_ch3:results_del_pop}. 

Results show that the synthetic population generated using ciDATGAN is generally more representative than the one generated using DATGAN. For the number of individuals per household and the household composition, ciDATGAN performs the best. This can be explained by the correlation between the individual's age and these two variables. Indeed, Figure~\ref{fig_ch3:corr_mat} shows a high negative correlation between age and the number of individuals per household. Thus, ciDATGAN learns this stronger correlation. Then, when we use the actual age distribution for the sampling phase, ciDATGAN can generate more representative individuals. For the other two variables, ciDATGAN performs similarly to the other two methodologies. Ethnicity is not linked to age in the DAG. Thus, correcting for the age, in this case, does not influence the ethnicity. Therefore, it explains that the results for the ethnicity did not improve between DATGAN and ciDATGAN. This link has been omitted deliberately as a control correlation to study the effect of the DAG in this process. Indeed, as shown in Figure~\ref{fig_ch3:corr_mat}, there is a positive correlation between age and ethnicity. It makes sense since migrant populations are more likely to be of working age. Thus, adding this link in the DAG, in Figure~\ref{fig_ch3:DAG_LTDS}, should improve the results as shown with the other variables. Finally, we cannot see any improvements in the number of cars/vans per household. It is expected since the correlation between age and the number of cars/vans per household is weak. It has, thus, been omitted in the DAG. Since the oversampled LTDS synthetic population shows similar results, adding this link would not improve the results of ciDATGAN.

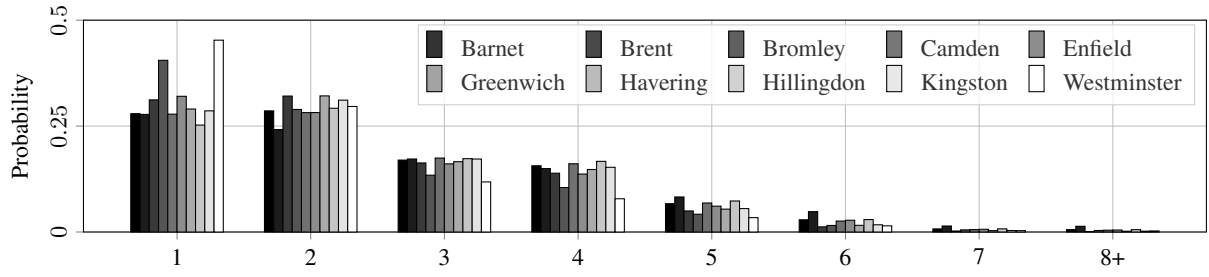
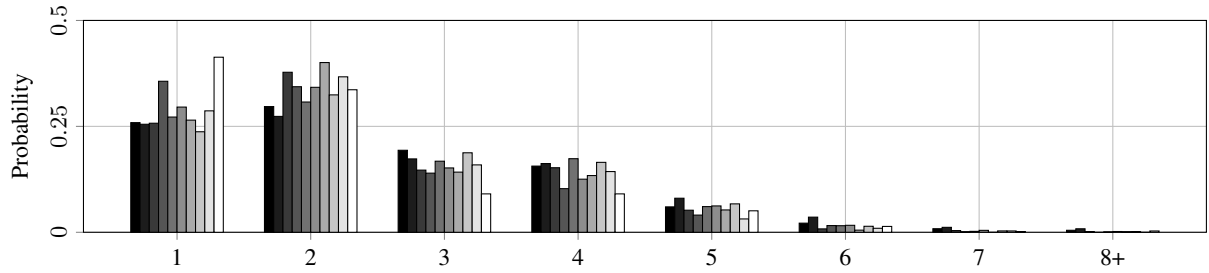
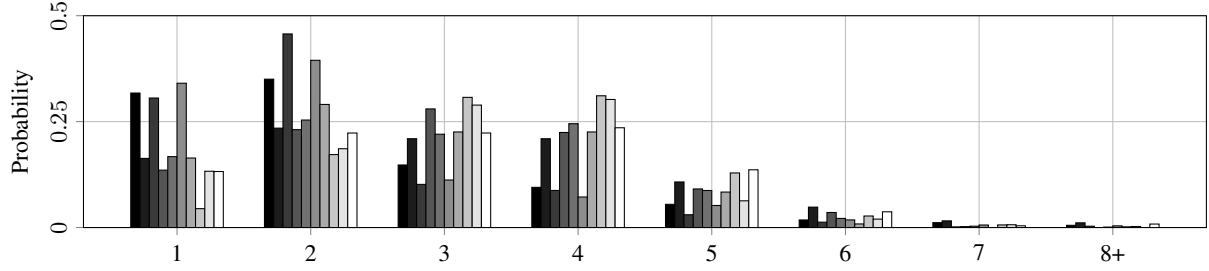
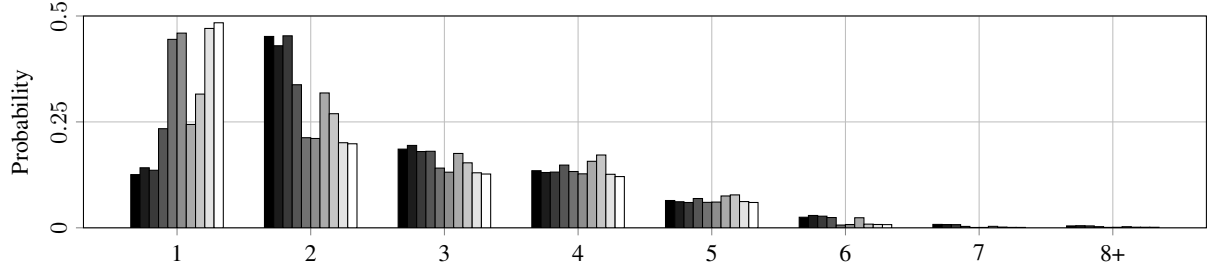
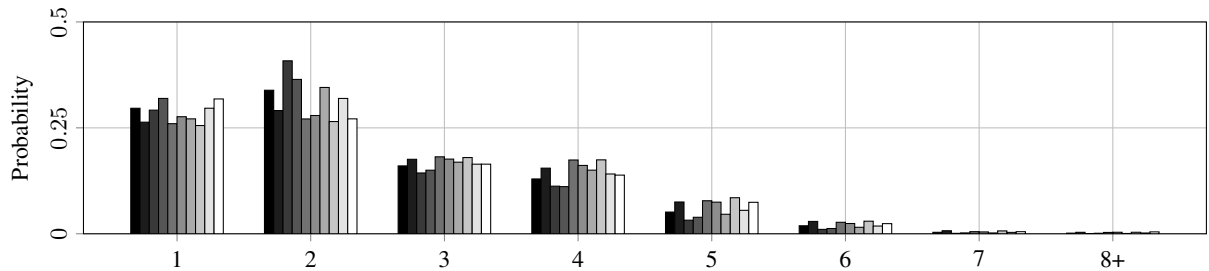
\begin{figure}[!htb]
    \centering
    \begin{subfigure}{\textwidth}
        \begin{tikzpicture}[inner frame sep=0]

    \begin{axis}[
        footnotesize,
        width=\textwidth,
        height=4.4cm,
        xtick={0,1,2,3,4,5,6,7},
        xticklabels={1, 2, 3, 4, 5, 6, 7, 8+},
        ytick={0, 0.25, 0.5},
        ylabel=Probability,
        every axis y label/.style={
            at={(ticklabel cs:0.5)},
            rotate=90,
            anchor=center,
            anchor=near ticklabel, 
            font=\footnotesize
        },
        yticklabel style={
                /pgf/number format/fixed,
                /pgf/number format/precision=2,
                rotate=90
        },
        ybar=0,
        bar width=3.5pt,
        legend columns=5,
        legend cell align={left},
        legend style={
            fill opacity=0.8, 
            draw opacity=1, 
            text opacity=1, 
            draw=white!80!black, 
            font=\footnotesize,
            /tikz/every even column/.append style={column sep=0.2cm},
            at={(0.99,0.98)},
            anchor=north east
        },
        tick align=outside,
        tick pos=left,
        xmajorgrids,
        ymajorgrids,
        legend image code/.code={\draw [#1] (0cm,-0.1cm) rectangle (0.2cm,0.25cm); },
        ymin=0,
        ymax=0.5,
    ]
    
    \foreach [evaluate=\i as \n using (\i-1)*100/(9)] \i in {1,2,...,10} {%
    	\edef\temp{\noexpand\addplot[draw=black,fill=white!\n!black] table[x index=0,y index=\i]{tikz/fig/nomis/size_nomis.dat};
    	}
    	\temp;
    }

    \legend{Barnet, Brent, Bromley, Camden, Enfield, Greenwich, Havering, Hillingdon, Kingston, Westminster}
    
    \end{axis}

\end{tikzpicture}
        \caption{Nomis data}
        \label{fig_ch3:size_nomis}
    \end{subfigure}
    \\
    \begin{subfigure}{\textwidth}
        \begin{tikzpicture}[inner frame sep=0]

    \begin{axis}[
        footnotesize,
        width=\textwidth,
        height=4.4cm,
        xtick={0,1,2,3,4,5,6,7},
        xticklabels={1, 2, 3, 4, 5, 6, 7, 8+},
        ytick={0, 0.25, 0.5},
        ylabel=Probability,
        every axis y label/.style={
            at={(ticklabel cs:0.5)},
            rotate=90,
            anchor=center,
            anchor=near ticklabel, 
            font=\footnotesize
        },
        yticklabel style={
                /pgf/number format/fixed,
                /pgf/number format/precision=2,
                rotate=90
        },
        ybar=0,
        bar width=3.5pt,
        legend columns=5,
        legend cell align={left},
        legend style={
            fill opacity=0.8, 
            draw opacity=1, 
            text opacity=1, 
            draw=white!80!black, 
            font=\footnotesize,
            /tikz/every even column/.append style={column sep=0.2cm},
            at={(0.99,0.98)},
            anchor=north east
        },
        tick align=outside,
        tick pos=left,
        xmajorgrids,
        ymajorgrids,
        legend image code/.code={\draw [#1] (0cm,-0.1cm) rectangle (0.2cm,0.25cm); },
        ymin=0,
        ymax=0.5
    ]
    
    \foreach [evaluate=\i as \n using (\i-1)*100/(9)] \i in {1,2,...,10} {%
    	\edef\temp{\noexpand\addplot[draw=black,fill=white!\n!black] table[x index=0,y index=\i]{tikz/fig/nomis/size_ltds.dat};
    	}
    	\temp;
    }

    
    \end{axis}

\end{tikzpicture}
        \caption{LTDS data}
        \label{fig_ch3:size_ltds}
    \end{subfigure}
    \\
    \begin{subfigure}{\textwidth}
        \begin{tikzpicture}[inner frame sep=0]

    \begin{axis}[
        footnotesize,
        width=\textwidth,
        height=4.4cm,
        xtick={0,1,2,3,4,5,6,7},
        xticklabels={1, 2, 3, 4, 5, 6, 7, 8+},
        ytick={0, 0.25, 0.5},
        ylabel=Probability,
        every axis y label/.style={
            at={(ticklabel cs:0.5)},
            rotate=90,
            anchor=center,
            anchor=near ticklabel, 
            font=\footnotesize
        },
        yticklabel style={
                /pgf/number format/fixed,
                /pgf/number format/precision=2,
                rotate=90
        },
        ybar=0,
        bar width=3.5pt,
        legend columns=5,
        legend cell align={left},
        legend style={
            fill opacity=0.8, 
            draw opacity=1, 
            text opacity=1, 
            draw=white!80!black, 
            font=\footnotesize,
            /tikz/every even column/.append style={column sep=0.2cm},
            at={(0.99,0.98)},
            anchor=north east
        },
        tick align=outside,
        tick pos=left,
        xmajorgrids,
        ymajorgrids,
        legend image code/.code={\draw [#1] (0cm,-0.1cm) rectangle (0.2cm,0.25cm); },
        ymin=0,
        ymax=0.5
    ]
    
    \foreach [evaluate=\i as \n using (\i-1)*100/(9)] \i in {1,2,...,10} {%
    	\edef\temp{\noexpand\addplot[draw=black,fill=white!\n!black] table[x index=0,y index=\i]{tikz/fig/nomis/size_ltds_bias.dat};
    	}
    	\temp;
    }

    
    \end{axis}

\end{tikzpicture}
        \caption{biased LTDS data}
        \label{fig_ch3:size_ltds_bias}
    \end{subfigure}
    \\
    \begin{subfigure}{\textwidth}
        \begin{tikzpicture}[inner frame sep=0]

    \begin{axis}[
        footnotesize,
        width=\textwidth,
        height=4.4cm,
        xtick={0,1,2,3,4,5,6,7},
        xticklabels={1, 2, 3, 4, 5, 6, 7, 8+},
        ytick={0, 0.25, 0.5},
        ylabel=Probability,
        every axis y label/.style={
            at={(ticklabel cs:0.5)},
            rotate=90,
            anchor=center,
            anchor=near ticklabel, 
            font=\footnotesize
        },
        yticklabel style={
                /pgf/number format/fixed,
                /pgf/number format/precision=2,
                rotate=90
        },
        ybar=0,
        bar width=3.5pt,
        legend columns=5,
        legend cell align={left},
        legend style={
            fill opacity=0.8, 
            draw opacity=1, 
            text opacity=1, 
            draw=white!80!black, 
            font=\footnotesize,
            /tikz/every even column/.append style={column sep=0.2cm},
            at={(0.99,0.98)},
            anchor=north east
        },
        tick align=outside,
        tick pos=left,
        xmajorgrids,
        ymajorgrids,
        legend image code/.code={\draw [#1] (0cm,-0.1cm) rectangle (0.2cm,0.25cm); },
        ymin=0,
        ymax=0.5
    ]
    
    \foreach [evaluate=\i as \n using (\i-1)*100/(9)] \i in {1,2,...,10} {%
    	\edef\temp{\noexpand\addplot[draw=black,fill=white!\n!black] table[x index=0,y index=\i]{tikz/fig/nomis/size_datgan.dat};
    	}
    	\temp;
    }

    
    \end{axis}

\end{tikzpicture}
        \caption{DATGAN, trained on biased LTDS data}
        \label{fig_ch3:size_datgan}
    \end{subfigure}
    \\
    \begin{subfigure}{\textwidth}
        \begin{tikzpicture}[inner frame sep=0]

    \begin{axis}[
        footnotesize,
        width=\textwidth,
        height=4.4cm,
        xtick={0,1,2,3,4,5,6,7},
        xticklabels={1, 2, 3, 4, 5, 6, 7, 8+},
        ytick={0, 0.25, 0.5},
        ylabel=Probability,
        every axis y label/.style={
            at={(ticklabel cs:0.5)},
            rotate=90,
            anchor=center,
            anchor=near ticklabel, 
            font=\footnotesize
        },
        yticklabel style={
                /pgf/number format/fixed,
                /pgf/number format/precision=2,
                rotate=90
        },
        ybar=0,
        bar width=3.5pt,
        legend columns=5,
        legend cell align={left},
        legend style={
            fill opacity=0.8, 
            draw opacity=1, 
            text opacity=1, 
            draw=white!80!black, 
            font=\footnotesize,
            /tikz/every even column/.append style={column sep=0.2cm},
            at={(0.99,0.98)},
            anchor=north east
        },
        tick align=outside,
        tick pos=left,
        xmajorgrids,
        ymajorgrids,
        legend image code/.code={\draw [#1] (0cm,-0.1cm) rectangle (0.2cm,0.25cm); },
        ymin=0,
        ymax=0.5
    ]
    
    \foreach [evaluate=\i as \n using (\i-1)*100/(9)] \i in {1,2,...,10} {%
    	\edef\temp{\noexpand\addplot[draw=black,fill=white!\n!black] table[x index=0,y index=\i]{tikz/fig/nomis/size_cidatgan.dat};
    	}
    	\temp;
    }

    
    \end{axis}

\end{tikzpicture}
        \caption{ciDATGAN, trained on biased LTDS data}
        \label{fig_ch3:size_cidatgan}
    \end{subfigure}
    \caption{Distributions of the number of individuals per household for each borough.}
    \label{fig_ch3:size_check}
\end{figure}

Lastly, we examine some distributions in more detail. Figure~\ref{fig_ch3:size_check} shows the distribution of individuals per household for each dataset. We see that the LTDS data (Figure~\ref{fig_ch3:size_ltds}) shows similar trends to the nomis data (Figure~\ref{fig_ch3:size_nomis}). This similarity was already shown in Figures~\ref{fig_ch3:ethn_check} and \ref{fig_ch3:cars_check}. On the other hand, the biased LTDS data (Figure~\ref{fig_ch3:size_ltds_bias}) shows different distributions with fewer households with a single individual. DATGAN can correct some boroughs as shown in Figure~\ref{fig_ch3:size_datgan}. However, depending on the borough, it mainly assigns one or two individuals per household. Finally, Figure~\ref{fig_ch3:size_cidatgan} shows that ciDATGAN is the best model for correcting the bias. Despite these results, we see that ciDATGAN cannot retrieve some of the trends, \emph{e.g.} boroughs near the center of London (Camden and Westminster) have more households with a single individual. This shows one of the limitations of ciDATGAN, \emph{i.e.} the model can only learn the logic in the original data. Thus, it will not be able to generate data that was never seen. Nonetheless, these results indicate that ciDATGAN has been able to address some of the bias in the LTDS dataset by using aggregate borough level control totals from the UK census. 

This final section shows that both DATGAN and ciDATGAN can generate large disaggregate synthetic populations. On an RTX 2080, both models were trained on the original LTDS dataset in around 25 minutes (1'000 epochs), and the sampling process took less than 10 minutes. This means that these models can generate a sizeable synthetic population in a short time. As shown in Figure~\ref{fig_ch3:size_check}, ciDATGAN can correct for possible biases in the feeder dataset using an unbiased distributor dataset. As the census data is fully aggregated at the borough level, there is no direct one-to-one correspondence between any individual in the synthetic population and the actual London population, maintaining privacy. However, the synthetic population, on an aggregate level, is representative of the London population. 

Creating synthetic populations for agent-based simulations can be laborious and imprecise, limiting their practical applications. However, ciDATGAN shows that one can create closely tailored and representative populations with limited human input in a short time.

\section{Conclusion}
\label{sec_ch3:conclusion}

This chapter presents the ciDATGAN model, an evolution of DATGAN that uses a novel type of conditionality for tabular data inspired by image completion. It consists of completing a large dataset with few variables (distributor dataset) using synthetic data learned from a smaller dataset with more variables (feeder dataset). We test this model on a trip-based dataset against DATGAN. First, we show that ciDATGAN provides equal or better quality data than the state-of-the-art DATGAN model. Then, we show that ciDATGAN, trained on a highly biased dataset, can perform as well as DATGAN trained on an unbiased dataset. It, thus, shows that ciDATGAN can be used to unbias a dataset if one has access to unbiased values for the conditional inputs. Finally, we show that ciDATGAN can learn the logic of the feeder dataset and can, therefore, be used to generate a large, detailed, and representative dataset using an external distributor dataset.

While ciDATGAN can generate unbiased and unknown datasets if good conditional inputs are provided, many research directions are still available. For example, ciDATGAN can only be used to generate independent rows, \emph{i.e.} cross-sectional data. A natural progression is, therefore, to allow for hierarchical data structures as in \cite{aemmer_generative_2022}, such as several individuals with correlated individual level attributes belonging to a single household with matching household level attributes. Complementary to generating hierarchical data, ciDATGAN could be upgraded to generate sequential data, such as activity patterns. However, the current architecture of the model cannot generate such data. Thus, one could combine ciDATGAN with another GAN designed to generate sequential data. For example, \cite{badu-marfo_composite_2020} developed a composite GAN that is composed of two different models: the first GAN generates the socioeconomic characteristics of the individuals and the second one generates sequential mobility data. However, in this case, the challenge is combining the methodology using the DAG with a new architecture that generates activity patterns.

\newpage
\bibliographystyle{plainnat-swapnames}
\bibliography{LedHilBie_ciDATGAN}
\newpage

\section{Appendix}

In the appendix, we present the two case studies presented in this article. 

\subsection{London Passenger Mode Choice (LPMC) dataset}

\begin{figure}[H]
    \centering
    \scriptsize
\begin{tikzpicture}[every text node part/.style={align=center}]


\def\ny{11}
\def\nx{8}

\def\dy{1.2}
\def\dx{3}
\def\shift{2pt}
\def\scale{0.8}

\tikzset{
block/.style={
	rounded corners=5pt,
    rectangle,
	fill=lightgray,
    draw=black,
	line width=1,
    text depth=.3\baselineskip, 
    text height=.7\baselineskip,
	inner ysep=0.2cm,
	inner xsep=0.2cm
},
doubleblock/.style={
	rounded corners=5pt,
    rectangle,
	fill=lightgray,
    draw=black,
	line width=1,
    text depth=.3\baselineskip, 
    text height=1.7\baselineskip,
	inner ysep=0.2cm,
	inner xsep=0.2cm
},
largeblock/.style={
	rounded corners=5pt,
    rectangle,
	fill=lightgray,
    draw=black,
	line width=1,
    text depth=.3\baselineskip, 
    text height=.7\baselineskip,
	minimum height=0.9cm,
	inner ysep=0.2cm,
	inner xsep=0.2cm
}}

\node[block, fill=lightred] (A) at (0,0) {\ttfamily distance};

\node[block, fill=lightgray] (B) at (\dx,2*\dy) {\ttfamily traffic\_percent};
\node[block, fill=lightgray] (C) at (1*\dx,1*\dy) {\ttfamily dur\_driving};
\node[block, fill=lightgray] (D) at (1*\dx,-1*\dy) {\ttfamily dur\_walking};
\node[block, fill=lightgray] (E) at (1*\dx,-2*\dy) {\ttfamily dur\_cycling};
\node[block, fill=lightgray] (F) at (1*\dx,-3*\dy) {\ttfamily dur\_pt};
\node[block, fill=lightblue] (G) at (1*\dx,-4*\dy) {\ttfamily hh\_region};

\node[block, fill=lightred] (H) at (3*\dx,2*\dy) {\ttfamily day\_of\_week};
\node[block, fill=lightred] (I) at (3*\dx,1*\dy) {\ttfamily start\_time};
\node[block, fill=lightred] (J) at (2*\dx,0*\dy) {\ttfamily travel\_mode};
\node[block, fill=lightblue] (K) at (2*\dx,-3*\dy) {\ttfamily hh\_vehicles};
\node[block, fill=lightblue] (L) at (2*\dx,-4*\dy) {\ttfamily hh\_income};

\node[block, fill=lightorange] (M) at (3*\dx,-2*\dy) {\ttfamily driving\_license};
\node[block, fill=lightblue] (N) at (3*\dx,-4*\dy) {\ttfamily hh\_people};

\node[block, fill=lightred] (O) at (4*\dx,2*\dy) {\ttfamily purpose};
\node[block, fill=lightorange] (P) at (4*\dx,-1*\dy) {\ttfamily fare\_type};
\node[block, fill=lightorange] (Q) at (4*\dx,-2*\dy) {\ttfamily age};
\node[block, fill=lightorange] (R) at (4*\dx,-3*\dy) {\ttfamily female};

\draw[-{Triangle[scale=\scale]}, line width=1] (H) -- (I);
\draw[-{Triangle[scale=\scale]}, line width=1] (H) -- (O);
\draw[-{Triangle[scale=\scale, color=white]}, line width=1] ([yshift=0.5*\shift]H.west) -- ([yshift=0.5*\shift]B.east);

\draw[-{Triangle[scale=\scale]}, line width=1] (B) -- (C);

\draw[-{Triangle[scale=\scale]}, line width=1] (L) -- (K);

\draw[-{Triangle[scale=\scale]}, line width=1] ([yshift=0.5*\shift]R.west) -| (M);
\draw[-{Triangle[scale=\scale]}, line width=1] ([yshift=-0.5*\shift]R.west) -| (N);

\coordinate (C0) at (3.6*\dx, -2*\dy);
\coordinate (C1) at (4.35*\dx, -2*\dy);

\draw[-{Triangle[scale=\scale]}, line width=1] ([yshift=0.5*\shift]Q.east) -- ([yshift=0.5*\shift]C1.west) |- (O);
\draw[-{Triangle[scale=\scale]}, line width=1] ([yshift=-0.5*\shift]Q.east) -- ([yshift=-0.5*\shift]C1.west) |- (N);
\draw[-{Triangle[scale=\scale]}, line width=1] (Q) -- (P);
\draw[-{Triangle[scale=\scale]}, line width=1] (Q) -- (M);
\draw[-{Triangle[scale=\scale, color=white]}, line width=1] ([yshift=1*\shift]Q.west) -- ([yshift=1*\shift]C0) |- ([yshift=-0.5*\shift, xshift=1*\shift]J.east);

\coordinate (C2) at (4*\dx, 2.5*\dy);

\draw[-{Triangle[scale=\scale]}, line width=1] (O) -- (C2) -| (A);

\coordinate (C3) at (2.5*\dx, -4*\dy);

\draw[-{Triangle[scale=\scale]}, line width=1] (N) -- (C3) |- (K);

\draw[-{Triangle[scale=\scale]}, line width=1] (K) -- (J);
\draw[-{Triangle[scale=\scale]}, line width=1] ([xshift=1*\shift]K.north) |- (M);

\coordinate (C4) at (1.5*\dx, -4*\dy);
\draw[-{Triangle[scale=\scale]}, line width=1] (G) -| (A);
\draw[-{Triangle[scale=\scale]}, line width=1] ([yshift=-1*\shift]G.east) -- ([yshift=-1*\shift]L.west);
\draw[-{Triangle[scale=\scale]}, line width=1] (G.east) -- (C4) |- (K);
\draw[-{Triangle[scale=\scale, color=white]}, line width=1] ([yshift=1*\shift]G.east) -- ([yshift=1*\shift, xshift=-1*\shift]C4.east) |- ([yshift=-0.5*\shift]J.west);

\coordinate (C5) at (0.5*\dx, 0);
\draw[-{Triangle[scale=\scale]}, line width=1] ([yshift=2.5*\shift]A.east) -- ([yshift=2.5*\shift, xshift=-1*\shift]C5) |- (B);
\draw[-{Triangle[scale=\scale]}, line width=1] ([yshift=1.5*\shift]A.east) -- ([yshift=1.5*\shift]C5) |- (C);
\draw[-{Triangle[scale=\scale, color=white]}, line width=1] ([yshift=0.5*\shift]A.east) -- ([yshift=0.5*\shift]J.west);
\draw[-{Triangle[scale=\scale]}, line width=1] ([yshift=-0.5*\shift]A.east) -- ([yshift=-0.5*\shift, xshift=1*\shift]C5) |- (D);
\draw[-{Triangle[scale=\scale]}, line width=1] ([yshift=-1.5*\shift]A.east) -- ([yshift=-1.5*\shift]C5) |- (E);
\draw[-{Triangle[scale=\scale]}, line width=1] ([yshift=-2.5*\shift]A.east) -- ([yshift=-2.5*\shift, xshift=-1*\shift]C5) |- (F);

\coordinate (C6) at (2*\dx, 1*\dy);
\draw[-{Triangle[scale=\scale, color=white]}, line width=1] (I) -- (C6) |- ([yshift=-0.5*\shift]B.east);

\draw[-{Triangle[scale=\scale]}, line width=1] ([xshift=-1*\shift]O.south) |- (I);
\draw[-{Triangle[scale=\scale, color=white]}, line width=1] (O.south) |- ([yshift=1.5*\shift, xshift=1*\shift]J.east);

\draw[-{Triangle[scale=\scale, color=white]}, line width=1] (P.north) |- ([yshift=0.5*\shift, xshift=1*\shift]J.east);

\draw[-{Triangle[scale=\scale, color=white]}, line width=1] (M.north) |- ([yshift=-1.5*\shift, xshift=1*\shift]J.east);

\draw[-{Triangle[scale=1.5*\scale]}, line width=1] ([xshift=0.5*\shift]B.east) -- (B.east);
\draw[-{Triangle[scale=1.5*\scale]}, line width=1] ([xshift=-0.5*\shift]J.west) -- (J.west);
\draw[-{Triangle[scale=2.4*\scale]}, line width=1] ([xshift=0.5*\shift]J.east) -- (J.east);
\end{tikzpicture}
    \caption{DAG used for the LPMC case study. Colors correspond to the category of variables: blue for households, orange for individuals, gray for alternatives, and red for trips. The conditional inputs are the variables \texttt{age}, \texttt{gender}, and \texttt{hh\_region}.}
    \label{fig_ch3:DAG_LPMC}
\end{figure}
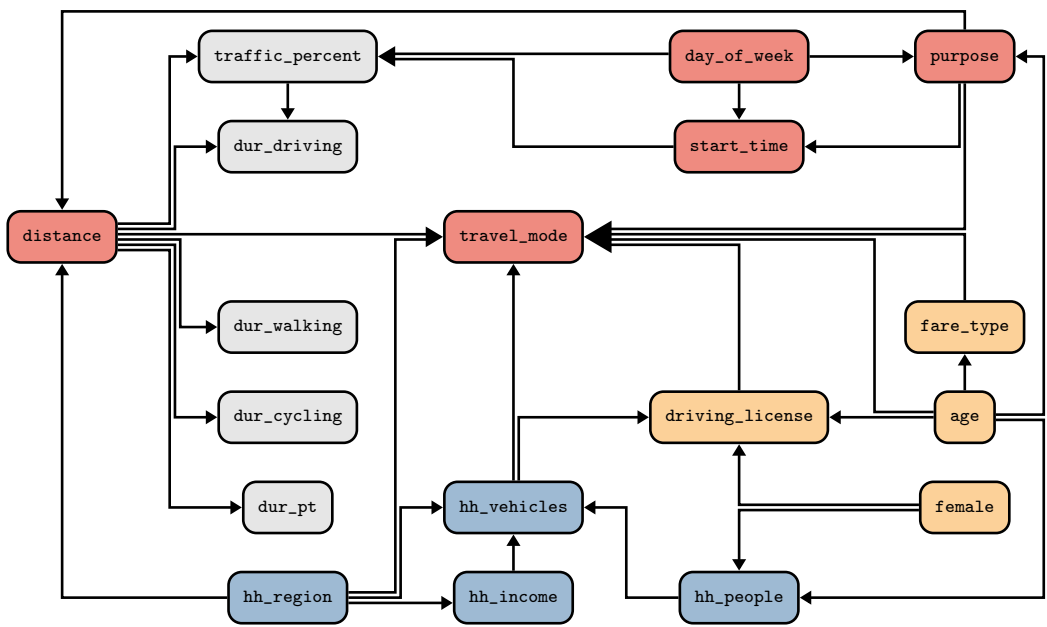

The LPMC~\citep{hillel_recreating_2018} dataset combines the LTDS records with matched trip trajectories and corresponding mode alternatives. The survey way conducted between April 2012 and March 2015 and records trips made by individuals residing and traveling within Greater London. The trip trajectories are extrapolated from Google Maps API. We selected a single trip per individual per household to avoid data leakage. The final dataset contains a total of 17'616 trips with 27 variables. The reader can contact the authors of \citep{hillel_recreating_2018} to get access to this dataset. The description of the variables is given in Table~\ref{tab_ch3:data_LPMC} and the associated DAG is given in Figure~\ref{fig_ch3:DAG_LPMC}.

\begin{xltabular}{\textwidth}{ccX}
    
    \caption{Details of the variables in the LPMC dataset. Colors correspond to the category of variables (in order): blue for households, orange for individuals, gray for alternatives, and red for trips. More details about these variables are given in~\cite{hillel_recreating_2018}.}
    \label{tab_ch3:data_LPMC} \\
    
    \toprule

    \textbf{Variables} & \textbf{Type} & \textbf{Description}  \\ \midrule
    \endfirsthead
    
    \multicolumn{3}{c}%
    {\tablename\ \thetable{} -- continued from previous page} \\
    \toprule
    \textbf{Variables} & \textbf{Type} & \textbf{Description}  \\ \midrule
    \endhead
    
    \botr
    \multicolumn{3}{r}{{Continues on next page...}}
    \endfoot
    
    \endlastfoot

    \bbox{\texttt{hh\_income}} & Categorical & Income of the household (10 categories) \\ 
    \bbox{\texttt{hh\_people}} & Categorical & Number of people in the household (11 values) \\ 
    \bbox{\texttt{hh\_region}} & Categorical & London region associated to the household (5 regions) \\ 
    \bbox{\texttt{hh\_vehicles}} & Categorical & Number of vehicles in the household (9 values)\\ \midr

    \obox{\texttt{age}} & Continuous & Age of individual in years \\ 
    \obox{\texttt{female}} & Categorical & Gender of the individual (0=male, 1=female) \\
    \obox{\texttt{driving\_license}} & Categorical & Whether the traveller has a driving licence (0=no, 1=yes)\\ 
    \obox{\texttt{fare\_type}} & Categorical & Public transport fare type of individual (5 categories)\\ \midr
    
    \gbox{\texttt{dur\_walking}} & Continuous & Duration of walking route\\ 
    \gbox{\texttt{dur\_cycling}} & Continuous & Duration of cycling route\\
    \gbox{\texttt{dur\_pt}} & Continuous & Duration of public transport route\\ 
    \gbox{\texttt{dur\_driving}} & Continuous & Duration of driving route\\ 
    \gbox{\texttt{traffic\_percent}} & Continuous & Traffic variability \\ \midr

    \rbox{\texttt{start\_time}} & Continuous & Start time of trip (in decimal hours)\\ 
    \rbox{\texttt{day\_of\_week}} & Categorical & Day of the week of travel (7 days)\\ 
    \rbox{\texttt{distance}} & Continuous & Straight line trip distance \\ 
    \rbox{\texttt{purpose}} & Categorical & Journey purpose for trip (5 categories)\\
    \rbox{\texttt{travel\_mode}} & Categorical & Mode of travel chosen by LTDS trip (4 categories)\\ 
    \botr
\end{xltabular}

\subsection{London Travel Data Survey (LTDS) dataset}

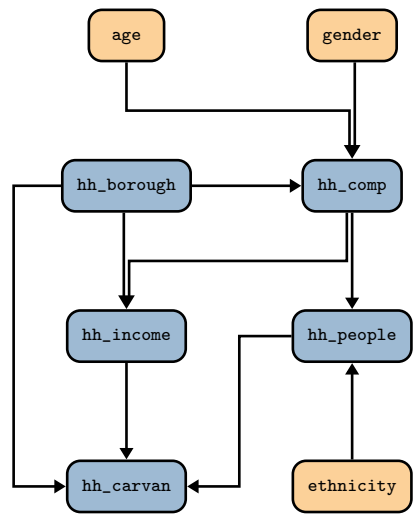
\begin{figure}[H]
    \centering
    \scriptsize
\begin{tikzpicture}[every text node part/.style={align=center}]

\def\ny{11}
\def\nx{8}

\def\dy{2}
\def\dx{3}
\def\shift{2pt}
\def\scale{0.8}

\tikzset{
block/.style={
	rounded corners=5pt,
    rectangle,
	fill=lightgray,
    draw=black,
	line width=1,
    text depth=.3\baselineskip, 
    text height=.7\baselineskip,
	inner ysep=0.2cm,
	inner xsep=0.2cm,
	minimum width=1cm,
}}

\node[block, fill=lightorange] (A) at (0,0) {\ttfamily age};
\node[block, fill=lightorange] (B) at (\dx,0) {\ttfamily gender};

\node[block, fill=lightblue] (C) at (0,-\dy) {\ttfamily hh\_borough};
\node[block, fill=lightblue] (D) at (\dx,-\dy) {\ttfamily hh\_comp};

\node[block, fill=lightblue] (E) at (0,-2*\dy) {\ttfamily hh\_income};
\node[block, fill=lightblue] (F) at (\dx,-2*\dy) {\ttfamily hh\_people};

\node[block, fill=lightblue] (H) at (0,-3*\dy) {\ttfamily hh\_carvan};
\node[block, fill=lightorange] (G) at (1*\dx,-3*\dy) {\ttfamily ethnicity};

\draw[-{Triangle[scale=\scale, color=white]}, line width=1] ([xshift=0.5*\shift]B.south) -- ([xshift=0.5*\shift]D.north);
\coordinate (C0) at (0, -0.5*\dy);
\draw[-{Triangle[scale=\scale, color=white]}, line width=1] (A) |- (C0) -| ([xshift=-0.5*\shift]D.north);
\draw[-{Triangle[scale=1.2*\scale]}, line width=1] ([yshift=\shift]D.north) -- (D.north);

\draw[-{Triangle[scale=\scale]}, line width=1] (C) -- (D);
\draw[-{Triangle[scale=\scale, color=white]}, line width=1] ([xshift=-0.5*\shift]C.south) -- ([xshift=-0.5*\shift]E.north);
\coordinate (C1) at (-0.5*\dx, -1*\dy);
\draw[-{Triangle[scale=\scale]}, line width=1] (C) -- (C1) |- (H);

\coordinate (C2) at (1*\dx, -1.5*\dy);
\draw[-{Triangle[scale=\scale, color=white]}, line width=1] ([xshift=-1*\shift]D.south) -- ([xshift=-1*\shift]C2) -| ([xshift=0.5*\shift]E.north);
\draw[-{Triangle[scale=1.2*\scale]}, line width=1] ([yshift=\shift]E.north) -- (E.north);
\draw[-{Triangle[scale=\scale]}, line width=1] (D) -- (F);

\draw[-{Triangle[scale=\scale]}, line width=1] (G) -- (F);

\draw[-{Triangle[scale=\scale]}, line width=1] (E) -- (H);

\coordinate (C3) at (0.5*\dx, -2.5*\dy);
\draw[-{Triangle[scale=\scale]}, line width=1] (F.west) -| (C3) |- (H);
\end{tikzpicture}
    \caption{DAG used for the LTDS case study. Colors correspond to the category of variables: blue for households and orange for individuals. The conditional inputs are the variables \texttt{age}, \texttt{gender}, and \texttt{hh\_borough}.}
    \label{fig_ch3:DAG_LTDS}
\end{figure}

The LTDS dataset contains unprocessed variables from the LPMC dataset. We only selected variables related to individuals and households. Some variables are available in the original LPMC dataset while other have been added from the collected surveys. The final dataset contains a total of 29'158 individuals and 8 variables. The description of the variables is given in Table~\ref{tab_ch3:data_LTDS} and the associated DAG is given in Figure~\ref{fig_ch3:DAG_LTDS}.

\begin{xltabular}{\textwidth}{ccX}
    \caption{Details of the variables in the LTDS dataset. Colors correspond to the category of variables (in order): blue for households and orange for individuals.}
    \label{tab_ch3:data_LTDS} \\
    
    \toprule

    \textbf{Variables} & \textbf{Type} & \textbf{Description}  \\ \midrule
    \endfirsthead
    
    \multicolumn{3}{c}%
    {\tablename\ \thetable{} -- continued from previous page} \\
    \toprule
    \textbf{Variables} & \textbf{Type} & \textbf{Description}  \\ \midrule
    \endhead
    
    \botr
    \multicolumn{3}{r}{{Continues on next page...}}
    \endfoot
    
    \endlastfoot

    \bbox{\texttt{hh\_income}} & Categorical & Income of the household (10 categories) \\ 
    \bbox{\texttt{hh\_people}} & Categorical & Number of people in the household (11 values) \\ 
    \bbox{\texttt{hh\_borough}} & Categorical & London borough associated to the household (10 boroughs) \\ 
    \bbox{\texttt{hh\_carvan}} & Categorical & Number of cars/vans in the household (8 values)\\ 
	\bbox{\texttt{hh\_comp}} & Categorical & Family composition in the household (4 categories)\\ \midr

    \obox{\texttt{age}} & Continuous & Age of individual in years \\ 
    \obox{\texttt{gender}} & Categorical & Gender of the individual (Male or Female) \\
    \obox{\texttt{ethnicity}} & Categorical & Ethnicity of the individual (5 categories)\\ 
    \botr
\end{xltabular}

\end{document}